\documentclass[11pt]{article}

\usepackage[final]{acl}

\usepackage{times}
\usepackage{latexsym}
\usepackage{enumitem}
\usepackage[T1]{fontenc}
\usepackage[utf8]{inputenc}

\usepackage{microtype}

\usepackage{inconsolata}

\usepackage{graphicx}

\usepackage{amsmath}
\usepackage{amssymb}
\usepackage{multirow}
\usepackage{hyperref}
\usepackage{kotex}
\usepackage[normalem]{ulem}
\useunder{\uline}{\ul}{}
\usepackage{color, colortbl}
\usepackage{booktabs}
\usepackage{adjustbox}
\usepackage{subcaption}
\usepackage{authblk}
\usepackage{url}

\usepackage{seqsplit}
\usepackage{longtable} 
\usepackage{array}
\usepackage{xcolor}    

\usepackage{pifont}  % X 기호용

\newcommand{\incr}[1]{#1}
\newcommand{\decr}[1]{#1}

\newcommand{\incu}{$\uparrow$}
\newcommand{\decd}{$\downarrow$}

\newcommand{\incck}{\textcolor{blue}{$\uparrow$}}   % 증가 예측 (정답)
\newcommand{\decck}{\textcolor{blue}{$\downarrow$}} % 감소 예측 (정답)
\newcommand{\incx}{\textcolor{red}{$\uparrow$}}     % 증가 예측 (오답)
\newcommand{\decx}{\textcolor{red}{$\downarrow$}}   % 감소 예측 (오답)

\definecolor{chembrayblue}{rgb}{0.6196, 0.7059 0.8275}
\definecolor{lightblue}{rgb}{0.886, 0.929, 0.996}
\definecolor{gray}{rgb}{0.937, 0.937, 0.937}
\definecolor{lp}{HTML}{C5C5E6}
\definecolor{my}{HTML}{E6F1FB}

\usepackage[most]{tcolorbox}

\definecolor{maincolor}{RGB}{0, 0, 0} 
\definecolor{subcolor}{RGB}{255, 255, 255}

\newtcolorbox{relationbox}[1]{
  colback=subcolor, % 본문 배경색
  colframe=gray, % 테두리 색
  fonttitle=\normalfont,
  coltitle=black,
  title={#1}, % 박스 제목
  sharp corners, % 둥근 모서리 싫으면 추가
  boxrule=0.5mm, % 테두리 두께
  left=2mm, right=2mm, top=1mm, bottom=1mm, % 여백
  enhanced, % 그림자 효과 등을 위해 필요
  drop shadow % 그림자 효과 (싫으면 삭제)
}

\title{MolSC: Leveraging Substituent Contributions to Enhance\\Fine-grained Molecular Understanding in LLMs}

\author{
  Hyuntae Park\textsuperscript{1} \quad
  Sooyeon Kim\textsuperscript{1} \quad
  Jiwon Park\textsuperscript{1} \quad
  SangKeun Lee\textsuperscript{1,2} \\
  \textsuperscript{1}Department of Artificial Intelligence, Korea University, Seoul, Republic of Korea \\
  \textsuperscript{2}Department of Computer Science and Engineering, Korea University, Seoul, Republic of Korea \\
  \texttt{\{pht0639, sooyeonkim, jiwonpark23, yalphy\}@korea.ac.kr} \\}

\begin{document}
\maketitle

\begin{abstract}
Recent advances in natural language processing have led to molecular Large Language Models (LLMs) with strong performance across diverse chemistry tasks.
However, they still struggle to capture fine-grained structure-property relationships, particularly how small, localized modifications alter a molecule's behavior.
To address this limitation, we introduce \texttt{MolSC}, a dataset of substituent contributions, defined as property changes induced by attaching specific substituents to molecular scaffolds.
Curated from manually annotated bioactivity records, \texttt{MolSC} spans structural-alert liability, target-specific bioactivity, and physicochemical descriptors, and contains 181K substituent-level examples for training.
We further propose \texttt{MolSC-Bench}, a held-out evaluation benchmark of 1,541 examples disjoint from \texttt{MolSC} at the scaffold, substituent, and molecule levels.
Our experiments show that existing molecular LLMs and strong proprietary models such as GPT-5.2 and Gemini-3-Flash show limited reliability in substituent contribution prediction.
In contrast, training on \texttt{MolSC} substantially improves this ability and achieves strong performance across diverse downstream molecular tasks.
These results highlight substituent contribution learning as a key component of fine-grained molecular understanding.\footnote{Our data and code are available at \url{https://github.com/Park-ing-lot/MolSC}.}
\end{abstract}

\section{Introduction}
Large Language Models (LLMs) have demonstrated remarkable capabilities across a wide range of natural language tasks. Building on this success, recent work has extended LLMs to the chemical domain by training them on textual molecular representations such as SMILES \citep{weininger1988smiles} and SELFIES \citep{krenn2020self} through molecular instruction tuning \citep{fang2023mol}. The resulting molecular LLMs \citep{m2024augmenting, instructmol} have achieved strong performance on diverse chemistry tasks, including reaction prediction \citep{fang2023mol}, molecular property prediction \citep{wu2018moleculenet}, and molecule-text translation \citep{edwards-etal-2022-translation}.

\begin{figure}[t]
    \centering
    \includegraphics[width=\linewidth]{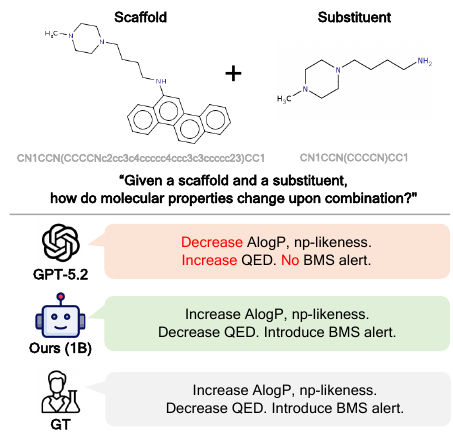}
    \caption{A \texttt{MolSC-Bench} example where our \texttt{MolSC}-trained model (Ours) correctly predicts property changes induced by substituent attachment, while GPT-5.2 predicts the opposite direction across all three property axes and misses the newly introduced BMS alert. Ground-truth (GT) changes are derived from ChEMBL.}
    \label{fig:fig1}
\end{figure}

% Despite recent advancements, current molecular LLMs struggle to capture fine-grained \textbf{structure-property relationships}, a fundamental ability underlying drug discovery tasks such as lead optimization and structure-property analysis \citep{free1964mathematical}. 
Despite recent advances, current molecular LLMs still struggle to capture fine-grained structure-property relationships, i.e., how local structural changes affect molecular properties and activities, a key ability for lead optimization in drug discovery \citep{free1964mathematical}.
Most existing molecular datasets pair each molecule with its overall property as a single unit \citep{wu2018moleculenet}, without supervising how individual substituents contribute to specific properties. Such molecule-level abstraction leads to representations that conflate functionally distinct compounds and restricts the model's ability to reason about local structural variations. Consequently, recent molecular LLMs struggle with functional group-level structure-property reasoning \citep{liu2026fgbench} and frequently misidentify substituents or misinterpret how they shape molecular properties \citep{yang2025knowmol}.

While a few studies have attempted to enhance such fine-grained molecular understanding by leveraging molecular fragments or functional groups \citep{yang2025knowmol, park2025bridging}, these approaches focus on aligning fragments with their textual descriptions or grouping them as structural tokens, without modeling the quantitative effect of each substituent on specific properties. As a result, how individual substituents shape specific molecular properties has remained largely implicit in existing datasets, motivating us to explicitly model substituent-level property contributions.

In this paper, we introduce \textbf{\texttt{MolSC}}, a dataset that captures how individual substituents contribute to molecular properties. We construct it by decomposing molecules from ChEMBL \citep{mendez2019chembl} into scaffolds and substituents, defining each substituent's contribution as the property difference between the molecule and its scaffold. These contributions span three axes: structural-alert liability, target-specific bioactivity, and physicochemical descriptors. In total, \texttt{MolSC} provides 181K substituent contributions as training signals, directly supervising how local structural changes shape molecular properties. We further propose \texttt{MolSC-Bench}, a held-out benchmark of 1,541 examples disjoint from the training set at the scaffold, substituent, and molecule levels, to evaluate how well molecular LLMs understand substituent contributions.

%  결과 보고 강조 필요
Using \texttt{MolSC-Bench}, we show that existing molecular LLMs and even powerful proprietary models such as GPT-5.2 and Gemini-3-Flash struggle to reliably predict substituent contributions.
In contrast, our model trained on \texttt{MolSC} with improved molecular representations not only achieves strong performance on \texttt{MolSC-Bench} and competitive results on downstream reaction-related and molecular property tasks. Notably, our 3B model surpasses prior approaches that rely on larger backbones and auxiliary 2D molecular graphs, using only 1D molecular representations.

Our main contributions are as follows:
\begin{itemize} \setlength\itemsep{0em}
    \item We introduce \texttt{MolSC} and \texttt{MolSC-Bench}, a training dataset and benchmark of measured substituent contributions.
    \item We reveal that existing molecular LLMs and powerful proprietary models such as GPT-5.2 often fail to predict substituent contributions.
    \item We show that training on \texttt{MolSC} improves downstream molecular tasks, highlighting the importance of substituent contributions.
\end{itemize}

\begin{figure*}[ht!]
    \centering
    \includegraphics[width=\textwidth]{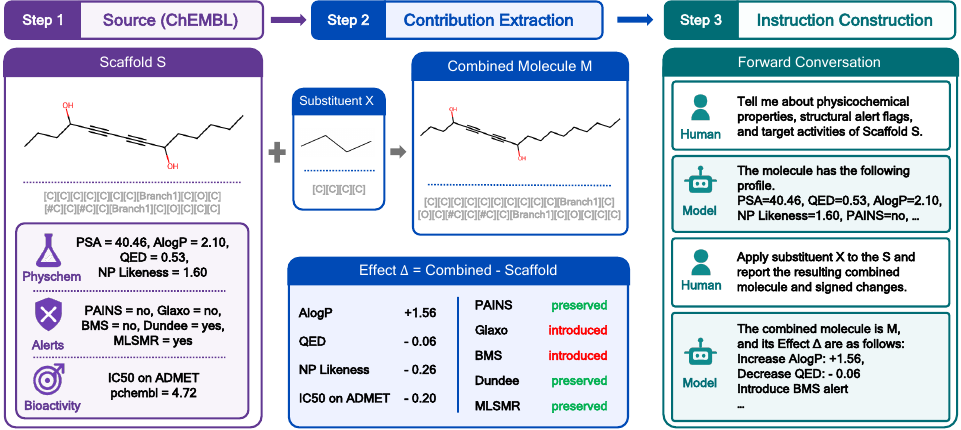}
    \caption{Data generation process of \texttt{MolSC}. We divide each molecule into a scaffold and a substituent, and define the substituent contribution as the property difference between the molecule and the scaffold. We then build an instruction-tuning dataset with predefined templates. Templates and examples are provided in Appendix~\ref{app:examples}.}
    \label{fig:main}
\end{figure*}

\section{Related Work}

\subsection{Molecule Language Modeling}
Driven by the success of deep learning and natural language processing, researchers have extensively applied these techniques to the chemical domain, interpreting molecules as either text sequences or atom-level graphs \citep{chithrananda2020chemberta, ross2022large, kim2024melt}. These approaches have gained attention for enabling more efficient drug candidate screening than traditional wet-lab pipelines.
More recently, molecule instruction-tuning methods \citep{fang2023mol, yu2024llasmol} have sought to leverage the adaptability of LLMs for molecular tasks. However, LLMs still exhibit limited structural understanding of molecules, motivating continued efforts to address this challenge \citep{park2024llamo, huomni}. For example, InstructMol integrates a 2D graph encoder with an LLM \citep{instructmol}, while subsequent work \citep{pei2025dmolt} further incorporates 3D molecular topology to improve generation reliability.
Nevertheless, existing approaches continue to struggle with fine-grained molecular understanding, often underperforming small-sized specialist models (<0.3B parameters) \citep{park2025bridging}. In this work, we aim to bridge this gap by explicitly modeling substituent contributions to molecular properties, enabling more precise molecular reasoning.

\subsection{Fine-grained Molecular Understanding}
Efforts to improve molecular understanding in language models have explored learning from fine-grained molecular decompositions. For property prediction, prior work has proposed fine-grained alignment across multiple molecular modalities \citep{feng2023unimap, yu2024multimodal, park2024moleco}, while molecule-caption alignment methods associate molecular fragments with textual phrases to capture structure-language correspondence \citep{zhang2025atomas, park2025bridging}.
To further enhance structural perception, hierarchical tokenization schemes encode atom-, motif-, and molecule-level information \citep{chenhierarchical}, and datasets with multi-level annotations have been introduced to better link molecular structures to textual descriptions \citep{yang2025knowmol}. In parallel, large-scale synthesis and decomposition pretraining has been shown to benefit reaction prediction tasks \citep{ijcai2025p0840}. However, these approaches primarily rely on surface-level signals, such as structural presence or alignment, without explicitly modeling how substructures influence task outcomes. In contrast, our work directly models how substituents change molecular properties across diverse chemical tasks.

\begin{table}[t]
\centering
\small
\begin{tabular}{lrr}
\toprule
 & \texttt{MolSC} & \texttt{MolSC-Bench} \\
\midrule
Contributions & 181,098 & 1,541 \\
\quad Unique scaffolds & 100,076 & 1,541 \\
\quad Unique substituents & 20,541 & 1,541 \\
\quad Unique molecules & 164,650 & 1,518 \\
Instructions & 149,232 & 1,541 \\
\bottomrule
\end{tabular}
\caption{Statistics of \texttt{MolSC} and \texttt{MolSC-Bench}. Additional statistics are in Appendix \ref{app:stats}.}
\label{tab:dataset_stats}
\end{table}

\section{Method}
In this section, we introduce our proposed dataset and training strategy. First, we describe the construction of \texttt{MolSC}, which provides explicit fragment contribution information that enables us to train and evaluate models on fine-grained structure-property signals (\S\ref{method:data}). Second, we describe how we leverage \texttt{MolSC} to train a molecular LLM that achieves deep molecular understanding (\S\ref{method:model}). Figure \ref{fig:main} illustrates the data construction process.

\subsection{\texttt{MolSC}}
\label{method:data}
The contribution of individual substituents to molecular properties has long been foundational in chemical modeling, where biological activity is often decomposed into additive substituent effects \citep{free1964mathematical}. This principle continues to underpin modern structure-activity and structure-property modeling \citep{cherkasov2014qsar, 10.1039/d0cs00098a}. However, such substituent-level information is rarely available to molecular LLMs as a direct learning signal, as existing datasets pair each molecule with its overall property and leave the effect of any individual substituent buried in molecule-level labels.

We construct \texttt{MolSC}, a dataset that explicitly captures substituent contributions to molecular properties. For each molecule formed by attaching a substituent to a scaffold, we record the signed property change induced by the substituent across three axes: structural-alert liability, target-specific bioactivity, and physicochemical descriptors.

\subsubsection{Substituent Contribution Extraction}

To obtain substituent-level supervision at scale, we ground \texttt{MolSC} in ChEMBL \citep{mendez2019chembl}, which provides annotations for all three axes central to drug discovery: structural-alert SMARTS patterns that flag liability concerns, assay-level bioactivity records that report target engagement, and physicochemical descriptors that summarize developability.

For each original molecule $M$ extracted from ChEMBL, we apply the Hussain-Rea fragmentation algorithm \citep{hussain2010computationally} to obtain scaffolds and substituents. This process severs a single acyclic bond between two heavy atoms, yielding an ordered pair $(S, X)$ where $S$ denotes the scaffold and $X$ the substituent. We retain only those scaffolds that themselves carry annotations along all three axes, ensuring that each scaffold provides a complete baseline against which the substituent's effect can be defined.

We then annotate substituent-induced changes along three property axes:
\begin{itemize}
\setlength\itemsep{0.2em}
    \item \textbf{Structural-alert liability.} Whether the substituent \textit{introduces}, \textit{removes}, or \textit{preserves} any alert from five expert-curated structural-alert sets: PAINS, Glaxo, BMS, Dundee, and MLSMR.
    
    \item \textbf{Target-specific bioactivity.} The signed activity change $\Delta_b = b_M - b_S$ for each \textit{(target, activity type)} pair measured for both the scaffold and the original molecule in the same assay, reducing inter-laboratory variability \citep{kalliokoski2013comparability}. We use eight standard ChEMBL activity types: IC50, Ki, Kd, EC50, AC50, XC50, Potency, and ED50.
    
    \item \textbf{Physicochemical descriptors.} The signed descriptor change $\Delta_P = P(M) - P(S)$ for five descriptors that can either increase or decrease after substituent attachment: PSA, ALogP, HBD, QED, and NP-likeness.\footnote{We exclude descriptors that change monotonically with attachment, such as molecular weight, heavy-atom count, and Lipinski violations.}
\end{itemize}
Together, these records yield approximately 1.2M substituent contributions, providing a substituent-resolved view of how local structural changes affect molecular properties.
Since we measure $\Delta_P$ for each (scaffold, substituent) pair, the same substituent has different $\Delta_P$ values depending on the scaffold.

\subsubsection{Data Refinement}
To improve the quality of the learning signal, we refine the extracted contributions by applying two filters. First, to ensure balanced learning across scaffolds, we drop over-represented scaffolds (those carrying more than ten distinct substituents, such as dinitrobenzene). Although they make up only 4.2\% of all scaffolds, they account for 40.8\% of the extracted contributions, and would otherwise dominate training at the expense of less frequent scaffolds. Second, we drop overly simple substituents with fewer than three heavy atoms, such as methyl and halogen groups, which act as atom-level decorations rather than chemically meaningful substituents. Single-atom substituents alone account for 42.5\% of all bioactivity records, and retaining them risks degenerate representations. These two filters remove approximately 78\% of the initial pool, yielding the final \texttt{MolSC} dataset comprising 181,098 substituent contributions.\footnote{We also exclude contributions whose $S$, $X$, or $M$ overlaps with any molecule in the downstream evaluation sets.}

\subsubsection{Instruction Construction}
To effectively leverage the instruction-following ability of LLMs for fine-grained structure-property reasoning, we verbalize each substituent contribution into a multi-turn instruction conversation. Each conversation is designed to supervise three abilities: profiling all molecular properties from ChEMBL, performing structural synthesis or decomposition, and predicting property changes when a specific substituent is attached. 

We construct conversations in both \textit{forward} and \textit{backward} directions, corresponding to substituent attachment and detachment, inspired by \citet{ijcai2025p0840}. These directions also reflect lead optimization and substructure analysis in drug discovery. In the forward direction, the model first receives $S$ and is asked to predict its full profile across the three axes; it is then given $X$ and asked to produce the resulting molecule $M$ together with the property changes induced by attaching $X$. In the backward direction, the model first receives $M$ and is asked to predict its full profile across the three axes; it is then asked to decompose $M$ into $S$ and $X$ together with the property contributions attributable to $X$. Contributions sharing the same anchor, either $S$ in the forward direction or $M$ in the backward direction, are merged into a single conversation so that the model can learn the substituent landscape of each anchor jointly. Examples and templates are provided in Appendix~\ref{app:examples}.

\begin{table*}[t]
\centering
\resizebox{\textwidth}{!}{%
\begin{tabular}{@{}l|cccc|ccccc@{}}
\toprule
\multirow{2}{*}[-0.5ex]{\textbf{Method}} & \multicolumn{4}{c|}{\textbf{Task 1: Property Profiling}} & \multicolumn{5}{c}{\textbf{Task 2: Substituent Contribution Prediction}} \\ \cmidrule(l){2-10} 
 & SR (\%) & propMAE $\downarrow$ & alertAcc $\uparrow$ & bioMAE $\downarrow$ & SR (\%) & $\Delta$propMAE $\downarrow$ & $\Delta$alertAcc $\uparrow$ & $\Delta$bioMAE $\downarrow$ & dirAcc $\uparrow$ \\ \midrule
Mol-Instruction-8B \citeyearpar{fang2023mol} & \phantom{0}40.6 & 24.019 & 0.667 & -- & \phantom{0}65.1 & 9.263 & 0.075 & \textbf{0.735} & 0.454 \\
ChemLLM-20B \citeyearpar{zhang2024chemllm} & \phantom{0}99.0 & 15.854 & 0.700 & 6.326 & 100.0 & 8.300 & 0.049 & 0.960 & 0.146 \\
ChemDFM-v2-14B \citeyearpar{zhao2025developing} & \phantom{0}98.8 & 5.627 & 0.627 & 7.312 & 100.0 & 7.928 & 0.068 & 0.804 & 0.199 \\
Gemini-3-Flash & 100.0 & 9.013 & 0.667 & 1.480 & 100.0 & 2.535 & 0.100 & 0.799 & 0.722 \\
GPT-5.2 & 100.0 & 10.531 & 0.517 & \textbf{0.657} & 100.0 & 3.519 & 0.219 & 2.471 & 0.706 \\ \midrule
\rowcolor{my} 
Ours-1B w/o SC, w/o dual-view & 100.0 & 0.616 & 0.947 & 0.817 & 100.0 & 9.305 & 0.422 & 0.776 & 0.525 \\
\rowcolor{my} 
Ours-1B w/o dual-view & 100.0 & 0.465 & 0.959 & 0.764 & 100.0 & 0.727 & 0.674 & 0.797 & 0.908 \\
\rowcolor{my} 
Ours-1B w/o SC & 100.0 & 0.404 & 0.963 & \underline{0.659} & \phantom{0}99.8 & 9.348 & 0.432 & 0.845 & 0.486 \\
\rowcolor{my} 
Ours-1B & 100.0 & \underline{0.386} & \underline{0.973} & 0.728 & 100.0 & \underline{0.639} & \underline{0.714} & \underline{0.754} & \underline{0.917} \\
\rowcolor{my} 
Ours-3B & 100.0 & \textbf{0.307} & \textbf{0.978} & 0.731 & 100.0 & \textbf{0.561} & \textbf{0.737} & \textbf{0.735} & \textbf{0.923} \\ \bottomrule
\end{tabular}%
}
\caption{Results on \texttt{MolSC-Bench}. SR denotes Success Rate, the proportion of outputs that follow the required format. SC denotes Substituent Contribution; w/o SC refers to a model trained only on property profiling from \texttt{MolSC}. dual-view refers to our training strategy described in \S\ref{method:model}. \textbf{Bold} and \underline{underline} indicate the best and second-best results, respectively. `--' denotes no valid response, making the metric unmeasurable.}
\label{tab:MolSC-Bench}
\end{table*}

\subsection{\texttt{MolSC-Bench}}
\label{method:bench}
For a clean evaluation of substituent-level reasoning, we construct \texttt{MolSC-Bench} from 1,541 contributions whose scaffold, substituent, and original molecule are all unseen in \texttt{MolSC}. From these contributions, we define two complementary evaluation tasks that probe molecular understanding at different levels of granularity. \textbf{Task~1 (Property Profiling)} provides a molecule and asks the model to predict its full profile across all three axes, evaluating whether the model can infer molecule-level properties. \textbf{Task~2 (Substituent Contribution Prediction)} provides a scaffold, a substituent to attach, and the scaffold's baseline values on the affected axes, and asks the model to predict the signed property changes induced by the attachment. Since the resulting molecule is never given, the model must infer the attachment outcome rather than read it off a known structure.

\subsection{Training Strategy}
\label{method:model}
We train a general-purpose LLM on \texttt{MolSC} using supervised fine-tuning. For a multi-turn conversation sequence $x = (x_1, \dots, x_T)$ constructed in \S\ref{method:data}, we optimize the standard causal language modeling objective:
$$\mathcal{L} = -\sum_{t=1}^{T} m_t \log p_\theta(x_t \mid x_{<t}),$$
where $m_t = 1$ for target tokens the model needs to generate, and $0$ otherwise.

To reliably learn local substituent contributions, the model first requires a coherent global molecular context. We establish this prerequisite through \textit{dual-view representation learning}, providing each molecule as both a SMILES and a SELFIES string in every turn. This joint formulation enforces structural consistency across complementary textual forms: SMILES \citep{weininger1988smiles} compactly exposes atom connectivity, whereas SELFIES \citep{krenn2020self} ensures chemical validity \citep{leon2024comparing}. By stabilizing the global representation, this dual-view alignment provides a robust basis for capturing fine-grained structure-property relationships.

\begin{table*}[t]
% \small
\centering
\resizebox{\textwidth}{!}{%
\begin{tabular}{l|c|c|ccccccc}
\toprule
\textbf{Method} & \# Params & Modality & Exact$\uparrow$ & BLEU$\uparrow$ & Lev$\downarrow$ & RDK$\uparrow$ & MAC$\uparrow$ & Mor$\uparrow$ & Validity$\uparrow$ \\ 
\midrule
\rowcolor{gray}
\multicolumn{10}{l}{\textit{Forward Reaction Prediction}} \\
Mol-Instructions \citep{fang2023mol} & 8B & 1D & 0.503 & 0.883 & 13.410 & 0.756 & 0.863 & 0.708 & 1.000 \\
InstructMol \citep{instructmol} & 7B & 1D + 2D & 0.536 & 0.967 & 10.851 & 0.776 & 0.878 & 0.741 & 1.000 \\
HIGHT \citep{chenhierarchical} & 7B & 1D + 2D & 0.037 & 0.869 & 23.759 & 0.590 & 0.394 & 0.340 & 0.993 \\
UniMoT \citep{guo2025unified} & 7B & 1D + 2D & 0.611 & 0.980 & 8.297 & 0.836 & 0.911 & 0.807 & 1.000 \\
Omni-Mol \citep{huomni} & 2B & 1D + 2D & 0.733 & 0.980 & 5.550 & 0.895 & 0.947 & 0.870 & 1.000 \\
KnowMol \citep{yang2025knowmol} & 7B & 1D + 2D & 0.752 & 0.986 & 5.662 & 0.889 & 0.943 & 0.877 & 1.000 \\
\rowcolor{my}
Ours-1B w/o SC, w/o dual-view & 1B & 1D & 0.762 & 0.985 & 5.311 & 0.895 & 0.946 & 0.878 & 1.000 \\
\rowcolor{my}
Ours-1B w/o dual-view & 1B & 1D & 0.790 & 0.986 & 4.916 & 0.917 & 0.955 & 0.899 & 1.000 \\
\rowcolor{my}
Ours-1B w/o SC & 1B & 1D & 0.926 & \uline{0.994} & \uline{1.228} & \uline{0.978} & \uline{0.990} & \uline{0.971} & 1.000 \\
\rowcolor{my}
Ours-1B & 1B & 1D & \uline{0.927} & 0.990 & 1.281 & 0.977 & 0.986 & 0.969 & 1.000 \\
\rowcolor{my}
Ours-3B & 3B & 1D & \textbf{0.948} & \textbf{0.995} & \textbf{0.760} & \textbf{0.984} & \textbf{0.992} & \textbf{0.979} & 1.000 \\ 
\midrule

\rowcolor{gray}
\multicolumn{10}{l}{\textit{Retrosynthesis}} \\
Mol-Instructions \citep{fang2023mol} & 8B & 1D & 0.333 & 0.842 & 17.642 & 0.704 & 0.815 & 0.646 & 1.000 \\
InstructMol \citep{instructmol} & 7B & 1D + 2D & 0.407 & 0.941 & 13.967 & 0.753 & 0.852 & 0.714 & 1.000 \\
HIGHT \citep{chenhierarchical} & 7B & 1D + 2D & 0.008 & 0.863 & 28.912 & 0.564 & 0.340 & 0.309 & 1.000 \\
UniMoT \citep{guo2025unified} & 7B & 1D + 2D & 0.478 & \uline{0.974} & 11.634 & 0.810 & 0.909 & 0.771 & 1.000 \\
Omni-Mol \citep{huomni} & 2B & 1D + 2D & 0.570 & 0.960 & 8.970 & 0.864 & 0.909 & 0.830 & 1.000 \\
KnowMol \citep{yang2025knowmol} & 7B & 1D + 2D & 0.598 & \textbf{0.975} & 8.363 & 0.856 & 0.912 & 0.829 & 1.000 \\ 
\rowcolor{my}
Ours-1B w/o SC, w/o dual-view & 1B & 1D & 0.581 & 0.960 & 9.031 & 0.854 & 0.907 & 0.823 & 1.000 \\
\rowcolor{my}
Ours-1B w/o dual-view & 1B & 1D & 0.595 & 0.963 & 8.077 & 0.876 & 0.919 & 0.840 & 1.000 \\
\rowcolor{my}
Ours-1B w/o SC & 1B & 1D & \uline{0.675} & 0.969 & 5.913 & 0.915 & \uline{0.941} & \uline{0.887} & 1.000 \\
\rowcolor{my}
Ours-1B & 1B & 1D & 0.674 & 0.971 & \uline{5.588} & \uline{0.915} & 0.940 & 0.885 & 1.000 \\
\rowcolor{my}
Ours-3B & 3B & 1D & \textbf{0.692} & 0.972 & \textbf{5.385} & \textbf{0.920} & \textbf{0.943} & \textbf{0.894} & 1.000 \\ \midrule

\rowcolor{gray}
\multicolumn{10}{l}{\textit{Reagent Prediction}} \\
Mol-Instructions \citep{fang2023mol} & 8B & 1D & 0.101 & 0.648 & 18.326 & 0.412 & 0.521 & 0.375 & 1.000 \\
InstructMol \citep{instructmol} & 7B & 1D + 2D & 0.129 & 0.610 & 19.664 & 0.444 & 0.539 & 0.400 & 1.000 \\
HIGHT \citep{chenhierarchical} & 7B & 1D + 2D & 0.050 & 0.462 & 28.970 & 0.441 & 0.314 & 0.275 & 1.000 \\
UniMoT \citep{guo2025unified} & 7B & 1D + 2D & 0.167 & 0.728 & 14.588 & \uline{0.549} & 0.621 & 0.507 & 1.000 \\
Omni-Mol \citep{huomni} & 2B & 1D + 2D & 0.230 & \uline{0.736} & 14.590 & \textbf{0.557} & \uline{0.627} & \uline{0.520} & 1.000 \\
KnowMol \citep{yang2025knowmol} & 7B & 1D + 2D & \uline{0.238} & 0.733 & \uline{14.058} & 0.525 & 0.609 & 0.490 & 1.000 \\
\rowcolor{my}
Ours-1B w/o SC, w/o dual-view & 1B & 1D & 0.201 & 0.690 & 16.363 & 0.484 & 0.584 & 0.458 & 1.000 \\
\rowcolor{my}
Ours-1B w/o dual-view & 1B & 1D & 0.206 & 0.702 & 15.752 & 0.497 & 0.592 & 0.469 & 1.000 \\
\rowcolor{my}
Ours-1B w/o SC & 1B & 1D & 0.230 & 0.726 & 14.617 & 0.506 & 0.605 & 0.489 & 1.000 \\
\rowcolor{my}
Ours-1B & 1B & 1D & 0.235 & 0.732 & 14.274 & 0.516 & 0.612 & 0.495 & 1.000 \\
\rowcolor{my}
Ours-3B & 3B & 1D & \textbf{0.262} & \textbf{0.747} & \textbf{13.456} & 0.548 & \textbf{0.636} & \textbf{0.529} & 1.000 \\

\bottomrule
\end{tabular}%
}
\caption{Results for reaction prediction tasks. \textbf{Bold} and \uline{underlined} mark the best and second-best scores. 1D and 2D refer to 1D molecular representations and 2D molecular graphs. }
\label{tab:main_results_mol_gen}
\end{table*}

\section{Experiments}
In this section, we demonstrate the efficacy of the substituent contribution information provided in \texttt{MolSC} through extensive experiments and analyses to answer the following questions:
\begin{itemize}\setlength\itemsep{0em}
    \item[\textbf{Q1}] Does \texttt{MolSC} offer a better understanding of substituent contribution? (\S\ref{molsc-bench})
    \item[\textbf{Q2}] Can the knowledge of \texttt{MolSC} be generalized to downstream tasks? (\S\ref{molecule generation}, \S\ref{property})
    \item[\textbf{Q3}] Does \texttt{MolSC} enable models to capture structure-property relationships? (\S\ref{main:analysis})
    % \item[\textbf{Q4}] Which factors in \texttt{MolSC} are crucial for capturing these relationships? (\S\ref{main:ablation})
\end{itemize}

\subsection{Experimental Settings}

\paragraph{Backbone.}
We use Llama-3.2-1B-Instruct and Llama-3.2-3B-Instruct \citep{grattafiori2024llama} as backbone models, and fine-tune them on \texttt{MolSC} with LoRA \citep{hu2022lora} using the training strategy described in Section~\ref{method:model}. Further implementation details are provided in Appendix~\ref{app:details}.

\paragraph{Dataset.}
For training and evaluating our models, we use \texttt{MolSC} and \texttt{MolSC-Bench}. We process all molecules using RDKit\footnote{\url{https://www.rdkit.org/}} and impose a maximum atom count of 100 due to computational constraints. For downstream tasks, we adopt the widely used benchmark, Mol-Instructions \citep{fang2023mol}. Specifically, among its diverse tasks, we select those relevant to our study: forward reaction prediction, retrosynthesis, reagent prediction, and molecular property regression. To prevent data leakage, we exclude all molecules appearing in the downstream validation and test sets from the \texttt{MolSC} training set. The performance of our backbone model on the downstream tasks is provided in Appendix \ref{app:backbone}. We additionally evaluate on BACE and BBBP from MoleculeNet \citep{wu2018moleculenet}, the Few subset of ACNet \citep{zhang2023activity}, and FGBench \citep{liu2026fgbench}.

\subsection{MolSC-Bench}
\label{molsc-bench}

\paragraph{Baselines.}
We compare our models against two groups of baselines. The first group consists of molecular LLMs, including Mol-Instruction-8B \citep{fang2023mol}, ChemLLM-20B \citep{zhang2024chemllm}, and ChemDFM-v2-14B \citep{zhao2025developing}. The second group consists of powerful proprietary models, GPT-5.2 and Gemini-3-Flash, accessed via their official APIs. All baselines receive the same prompts as shown in Figures \ref{appendix:test-task1} and \ref{appendix:test-task2}.

\paragraph{Metrics.}
Across both tasks, we first report the Success Rate (SR), the proportion of outputs that adhere to the required format. To evaluate the model's predictions, we measure the Mean Absolute Error (MAE) for the physicochemical and bioactivity axes, and prediction accuracy for the structural-alert axis.
Specifically, in Task 1, we evaluate the absolute profile of the input molecule. Thus, the MAE is computed between the predicted and actual property values (propMAE, bioMAE), and the accuracy is calculated over all alert flags (alertAcc). In Task 2, we evaluate the property changes induced by substituent attachment. Accordingly, the MAE is computed against the ground-truth property changes ($\Delta$propMAE, $\Delta$bioMAE), and the accuracy is measured exclusively over the modified alert flags ($\Delta$alertAcc). Additionally, Task 2 includes direction accuracy (dirAcc) to report the proportion of correctly predicted change directions.

\begin{table}[t!]
\footnotesize
\centering
\resizebox{\columnwidth}{!}{%
\begin{tabular}{@{}l|cccc@{}}
\toprule
\textbf{Method} & HOMO$\downarrow$ & LUMO$\downarrow$ & $\Delta\epsilon\downarrow$ & Avg$\downarrow$ \\ \midrule
Mol-Instruction \citeyearpar{fang2023mol} & 0.0210 & 0.0210 & 0.0203 & 0.0210 \\
InstructMol \citeyearpar{instructmol} & 0.0048 & 0.0050 & 0.0061 & 0.0050 \\
HIGHT \citeyearpar{chenhierarchical} & 0.0056 & 0.0065 & 0.0077 & 0.0066 \\
UniMoT \citeyearpar{guo2025unified} & 0.0042 & 0.0047 & 0.0055 & 0.0049 \\
Omni-Mol \citeyearpar{huomni} & 0.0038 & 0.0047 & 0.0049 & 0.0044 \\
KnowMol \citeyearpar{yang2025knowmol} & 0.0031 & 0.0037 & 0.0035 & 0.0034 \\
\rowcolor{my}
Ours w/o SC, w/o dual-view & 0.0033 & 0.0038 & 0.0040 & 0.0037 \\
\rowcolor{my}
Ours w/o dual-view & 0.0034 & 0.0035 & 0.0040 & 0.0036 \\
\rowcolor{my}
Ours w/o SC & \textbf{0.0027} & \textbf{0.0027} & \textbf{0.0032} & \textbf{0.0029} \\
\rowcolor{my}
Ours-1B & \uline{0.0028} & \uline{0.0030} & \uline{0.0034} & \uline{0.0030} \\
\bottomrule
\end{tabular}%
}
\caption{Results for molecular property regression tasks on the QM9 benchmark \citep{fang2023mol} evaluated using Mean Absolute Error (MAE). Results for our variants are averaged over three random seeds. \textbf{Bold} and \uline{underline} indicate the best and second-best results.}
\label{tab:property_reg}
\end{table}

% \begin{table}[t!]
% \centering
% \small
% \resizebox{\columnwidth}{!}{%
% \begin{tabular}{@{}l|cccc@{}}
% \toprule
% \textbf{Method} & BACE$\uparrow$ & BBBP$\uparrow$ & ACNet$\uparrow$ & Avg$\uparrow$ \\
% \midrule
% Llama-3.2-1B-Instruct & 83.50 & 71.00 & 74.12 & 76.21 \\
% \rowcolor{my}
% Ours w/o SC, w/o dual-view & 85.98 & \textbf{73.12} & 77.32 & 78.81 \\
% \rowcolor{my}
% Ours w/o dual-view & \textbf{88.59} & 70.71 & \uline{77.93} & \uline{79.08} \\
% \rowcolor{my}
% Ours w/o SC & 82.42 & 70.81 & 77.30 & 76.84 \\
% \rowcolor{my}
% Ours-1B & \uline{86.58} & \uline{72.01} & \textbf{79.89} & \textbf{79.49} \\
% \bottomrule
% \end{tabular}
% }
% \caption{Results for drug-like molecular property tasks evaluated using ROC-AUC. BACE and BBBP are from MoleculeNet \citep{wu2018moleculenet}. \textbf{Bold} and \uline{underline} indicate the best and second-best results.}
% \label{tab:druglike}
% \end{table}

\paragraph{Results.}
Models trained on \texttt{MolSC} outperform all baselines on \texttt{MolSC-Bench}, as shown in Table~\ref{tab:MolSC-Bench}. While proprietary models perform better than existing molecular LLMs on Task~1, both groups perform poorly on Task~2, indicating that substituent contribution prediction remains difficult even for stronger general-purpose models. In particular, baselines often predict plausible magnitudes but fail to capture the correct direction of change: molecular LLMs are near chance, and proprietary models reach only around $0.7$ direction accuracy. In contrast, our 3B model achieves the best overall performance, with a propMAE of $0.307$ on Task~1 and a direction accuracy of $0.923$ on Task~2. Since \texttt{MolSC} and \texttt{MolSC-Bench} share no scaffolds, substituents, or original molecules, these results suggest that \texttt{MolSC} teaches generalizable substituent contribution prediction.

\paragraph{Ablation Study.}
We ablate the Substituent Contribution supervision (SC) and dual-view representation learning introduced in \S\ref{method:model}. Without SC, the model trained only on property profiling performs poorly on Task~2, showing that molecule-level property supervision alone is insufficient for substituent contribution prediction. Adding SC substantially improves Task~2 in both settings ($0.908$ and $0.917$), while dual-view mainly benefits Task~1 ($0.616 \rightarrow 0.404$). Finally, dual-view representation learning yields the best overall performance, indicating that it further helps fine-grained structure-property understanding.

\begin{table}[t!]
\centering
\small
\resizebox{\columnwidth}{!}{%
\begin{tabular}{@{}l|cccc@{}}
\toprule
\textbf{Method} & BACE$\uparrow$ & BBBP$\uparrow$ & ACNet$\uparrow$ & Avg$\uparrow$ \\
\midrule
Llama-3.2-1B-Instruct & 83.50 & 71.00 & 74.12 & 76.21 \\
\rowcolor{my}
Ours w/o SC, w/o dual-view & 85.98 & \textbf{73.12} & 77.32 & 78.81 \\
\rowcolor{my}
Ours w/o dual-view & \textbf{88.59} & 70.71 & \uline{77.93} & \uline{79.08} \\
\rowcolor{my}
Ours w/o SC & 82.42 & 70.81 & 77.30 & 76.84 \\
\rowcolor{my}
Ours-1B & \uline{86.58} & \uline{72.01} & \textbf{79.89} & \textbf{79.49} \\
\bottomrule
\end{tabular}
}
\caption{Results for drug-like molecular property tasks evaluated using ROC-AUC. BACE and BBBP are from MoleculeNet \citep{wu2018moleculenet}. \textbf{Bold} and \uline{underline} indicate the best and second-best results.}
\label{tab:druglike}
\end{table}

\subsection{Molecular Reaction Prediction}
\label{molecule generation}

\paragraph{Baselines.}
We compare our model with a diverse set of competitive Molecule LLMs, including the model from Mol-Instructions \citep{fang2023mol}, InstructMol \citep{instructmol}, HIGHT \citep{chenhierarchical}, UniMoT \citep{guo2025unified}, Omni-Mol \citep{huomni}, and KnowMol \citep{yang2025knowmol}. In particular, we focus on comparisons with Omni-Mol, which shares the same backbone language model as our approach (Llama-3.2-1B-Instruct \citep{grattafiori2024llama}).

\paragraph{Metrics.}
We employ Exact Match (EM), BLEU-4 \citep{bleu}, and Levenshtein distance (Lev) \citep{levenshtein1966binary} to measure string-level similarity, along with Validity to assess the grammatical correctness of generated molecules. We also adopt molecular fingerprint-based similarity metrics, including MACCS FTS (MAC) \citep{DBLP:journals/jcisd/DurantLHN02}, RDK FTS (RDK) \citep{DBLP:journals/jcisd/SchneiderSL15}, and Morgan FTS (Mor) \citep{DBLP:journals/jcisd/RogersH10}, to compare structural similarity with reference molecules. All evaluations are conducted using model-generated SELFIES representations.

\paragraph{Results.}
Models trained on \texttt{MolSC} demonstrate strong overall performance across reaction-related tasks, including forward reaction, retrosynthesis, and reagent prediction (Table~\ref{tab:main_results_mol_gen}). Notably, in forward reaction and retrosynthesis, our 1B model achieves an average Exact Match improvement of 14.9\%p over Omni-Mol, while Ours-3B achieves the highest performance across all three tasks. Since molecular properties are known to be related to chemical reactivity \citep{ahneman2018predicting, sandfort2020structure}, we interpret the substituent-level structure-property knowledge in \texttt{MolSC} as one factor contributing to these gains. Furthermore, our models achieve these results relying solely on 1D representations, demonstrating that the structural knowledge in \texttt{MolSC} effectively transfers to downstream tasks without requiring complex 2D graph encoders or larger backbones.

\paragraph{Ablation Study.}
Without dual-view, adding SC improves Exact Match across all three tasks, from $0.762$ to $0.790$ on forward reaction prediction, $0.581$ to $0.595$ on retrosynthesis, and $0.201$ to $0.206$ on reagent prediction. Dual-view representation learning yields further substantial gains, reaching $0.927$, $0.674$, and $0.235$, respectively. Moreover, compared with the synthesis/decomposition-only variant, the full supervision improves forward reaction prediction from $0.917$ to $0.927$ and retrosynthesis from $0.658$ to $0.674$. These results demonstrate that substituent-contribution supervision provides useful additional signals for reaction-related tasks.

\subsection{Molecular Property Prediction}
\label{property}
\paragraph{Baselines and Metrics.}
We use the same baselines as in Section~\ref{molecule generation}, and report MAE on the QM9 regression tasks of Mol-Instructions \citep{fang2023mol}. Since QM9 consists of small organic molecules, we additionally evaluate on three drug-like benchmarks, BACE and BBBP from MoleculeNet \citep{wu2018moleculenet} and ACNet \citep{zhang2023activity}, and report ROC-AUC.

\paragraph{Results.}
As shown in Table 4, our 1B variants outperform the prior molecular LLM baselines on QM9. In particular, the model trained without SC but with dual-view representation learning achieves the best average MAE of 0.00287. Across three random seeds, dual-view substantially improves the average MAE from 0.00370 to 0.00287 without SC (\(p=0.010\)), whereas SC provides only a limited standalone benefit. These results indicate that the QM9 improvement primarily stems from dual-view representation learning.

\paragraph{Ablation Study.}
Unlike the previous tasks, property prediction concerns the properties of the whole input molecule, where property profiling plays a larger role than substituent contribution supervision. Indeed, on QM9 the model trained only on property profiling already performs well, and adding SC does not improve it further. The gain on QM9 therefore comes from the dual-view representation rather than from substituent-level knowledge, indicating that a robust molecular representation is important for property prediction. On the drug-like tasks, however, the two components reverse their roles: adding SC consistently improves the average, whereas the dual-view representation is marginal and in some cases degrades performance.

\begin{table}[t]
\centering
\small
\resizebox{\columnwidth}{!}{%
\begin{tabular}{@{}l|ccc@{}}
\toprule
\textbf{Model} & Single$\uparrow$ & Interaction$\uparrow$ & Comparison$\uparrow$ \\
\midrule
Mol-Instructions-8B \citeyearpar{fang2023mol} & 0.107 & 0.059 & 0.469 \\
ChemLLM-7B \citeyearpar{zhang2024chemllm} & 0.233 & 0.235 & 0.250 \\
LlaSMol-Mistral-7B \citeyearpar{yu2024llasmol} & 0.387 & 0.298 & 0.239 \\
Qwen2.5-7B \citeyearpar{qwen2025qwen25technicalreport}& 0.590 & 0.396 & 0.664 \\
nach0-base \citeyearpar{livne2024nach0} & 0.606 & 0.543 & 0.041 \\
Llama-3.2-1B-Instruct \citeyearpar{grattafiori2024llama} & 0.677 & 0.655 & 0.662 \\
Llama-3.1-70B \citeyearpar{grattafiori2024llama} & 0.683 & 0.530 & 0.456 \\
GPT-4o & 0.667 & 0.488 & 0.614 \\
o3-mini & \uline{0.687} & \uline{0.693} & \uline{0.703} \\
\rowcolor{my}
Ours-1B & \textbf{0.785} & \textbf{0.780} & \textbf{0.776} \\
\bottomrule
\end{tabular}
}
\caption{Results for functional group-level property reasoning on FGBench \citep{liu2026fgbench}, evaluated using boolean accuracy. \textbf{Bold} and \uline{underline} indicate the best and second-best results.}
\label{tab:fgbench}
\end{table}

\subsection{Functional Group-Level Property Reasoning}
\label{main:fgbench}

To examine whether the substituent contributions learned from \texttt{MolSC} transfer to other forms of fine-grained molecular understanding, we evaluate our model on FGBench \citep{liu2026fgbench}, which asks how one or more functional groups affect a specific molecular property.
The benchmark covers three question categories: the impact of a single functional group, the interaction of multiple functional groups within a molecule, and the direct comparison of two molecules that differ in their functional groups.

As shown in Table~\ref{tab:fgbench}, \texttt{Ours-1B} outperforms its backbone across all three dimensions, and achieves higher accuracy than baselines including Llama-3.1-70B and strong proprietary models such as GPT-4o and o3-mini. This shows that training on \texttt{MolSC} substantially improves the understanding of functional group-level property effects.

\subsection{Context-Dependent Contribution Analysis}
\label{main:analysis}

To examine whether \texttt{MolSC} enables models to capture scaffold-dependent structure-property relationships, we analyze cases where the same substituent induces opposite property changes depending on the scaffold. These cases require modeling the interaction between the substituent and the scaffold, rather than the substituent alone. We attach morpholine to four scaffolds for which the ground-truth effect on \texttt{qed} is positive for two scaffolds and negative for the other two, and compare \texttt{Ours-3B} with GPT-5.2 in Table~\ref{fig:context_dependent}. These records are drawn from contributions excluded from \texttt{MolSC} during construction due to overlap with downstream molecules, and are therefore unseen during \texttt{MolSC} training.

The two models show a clear difference. On axes with scaffold-independent directions, such as \texttt{psa} and \texttt{alogp}, both models predict the correct direction across all four scaffolds, suggesting that they can capture simpler substituent-driven effects. The difference emerges on the context-dependent \texttt{qed} axis. GPT-5.2 predicts the wrong \texttt{qed} direction for all four scaffolds, failing to capture how the scaffold changes the effect of the same substituent. In contrast, \texttt{Ours-3B} correctly tracks the sign change across all four cases, recognizing that morpholine can either increase or decrease \texttt{qed} depending on where it is attached. This suggests that training on \texttt{MolSC} helps the model learn scaffold-dependent substituent contributions, rather than relying only on a fixed substituent-level prior.

To extend this analysis beyond a single substituent, we draw from the same excluded contributions and select scaffolds that allow pair comparison, constructing $928$ combinations in which each of $464$ substituents is attached to two different scaffolds, and evaluate both models against the ground-truth direction of property change. As shown in Table~\ref{tab:context928}, \texttt{Ours-1B} substantially outperforms GPT-5.2 in direction accuracy, and the gap widens on the $359$ conflict-pairs, where the same substituent acts in opposite directions depending on the scaffold ($0.582$ versus $0.145$). This shows that the morpholine case above holds at scale rather than in a few isolated cases.

\begin{table}[t]
\centering
\large
\renewcommand{\arraystretch}{1.15}
\resizebox{\columnwidth}{!}{%
\begin{tabular}{@{}cl rr rr rr@{}}
% === 이 부분을 새로 추가합니다 ===
\multicolumn{8}{c}{\textbf{Task:} Predict property $\Delta$ upon adding morpholine (\raisebox{-0.3\height}{\includegraphics[width=1.0cm]{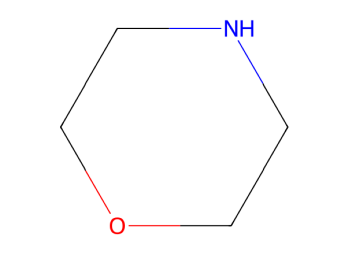}})} \\
\toprule
% =================================
& & \multicolumn{2}{c}{\textbf{GT $\Delta$}} & \multicolumn{2}{c}{\textbf{Ours-3B}} & \multicolumn{2}{c}{\textbf{GPT-5.2}} \\
\cmidrule(lr){3-4}\cmidrule(lr){5-6}\cmidrule(lr){7-8}
\textbf{Scaffold} & \textbf{Axis} & $\Delta$ & dir & $\Delta$ & dir & $\Delta$ & dir \\
\midrule

\multirow{4}{*}{\includegraphics[width=2.3cm]{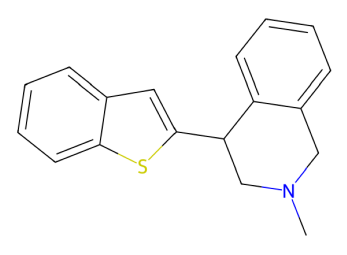}}
 & psa          & \incr{+12.47} & \incu & \incr{+12.47} & \incck & \incr{+28.96} & \incck \\
 & alogp        & \decr{-0.16}  & \decd & \decr{-0.34}  & \decck & \decr{-1.28}  & \decck \\
 & qed          & \incr{+0.04}  & \incu & \incr{+0.04}  & \incck & \decr{-0.02}  & \decx  \\
 & np\_likeness & \decr{-0.34}  & \decd & \decr{-0.49}  & \decck & \decr{-0.12}  & \decck \\
\midrule

\multirow{4}{*}{\includegraphics[width=2.3cm]{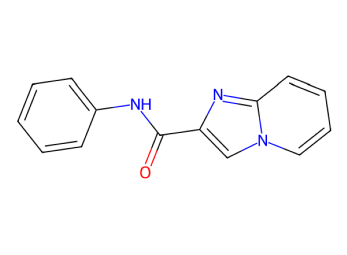}}
 & psa          & \incr{+12.47} & \incu & \incr{+12.47} & \incck & \incr{+32.80} & \incck \\
 & alogp        & \decr{-0.17}  & \decd & \decr{-0.08}  & \decck & \decr{-0.87}  & \decck \\
 & qed          & \incr{+0.06}  & \incu & \incr{+0.06}  & \incck & \decr{-0.06}  & \decx  \\
 & np\_likeness & \decr{-0.09}  & \decd & \incr{+0.01}  & \incx  & \incr{+0.74}  & \incx  \\
\midrule

\multirow{4}{*}{\includegraphics[width=2.3cm]{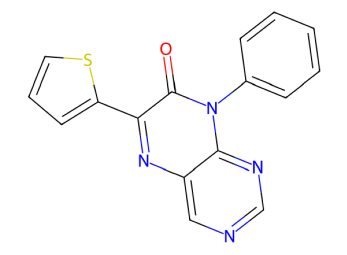}}
 & psa          & \incr{+12.47} & \incu & \incr{+12.47} & \incck & \incr{+27.83} & \incck \\
 & alogp        & \decr{-0.16}  & \decd & \decr{-0.08}  & \decck & \decr{-0.80}  & \decck \\
 & qed          & \decr{-0.04}  & \decd & \decr{-0.03}  & \decck & \incr{+0.04}  & \incx  \\
 & np\_likeness & \decr{-0.15}  & \decd & \decr{-0.14}  & \decck & \incr{+0.43}  & \incx  \\
\midrule

\multirow{4}{*}{\includegraphics[width=2.3cm]{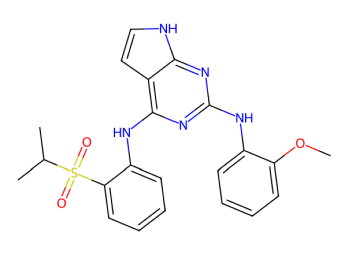}}
 & psa          & \incr{+12.47} & \incu & \incr{+12.47} & \incck & \incr{+25.00} & \incck \\
 & alogp        & \decr{-0.17}  & \decd & \decr{-0.08}  & \decck & \decr{-0.84}  & \decck \\
 & qed          & \decr{-0.08}  & \decd & \decr{-0.08}  & \decck & \incr{+0.06}  & \incx  \\
 & np\_likeness & \decr{-0.36}  & \decd & \decr{-0.28}  & \decck & \incr{+0.23}  & \incx  \\
\bottomrule
\end{tabular}
}
\caption{Property changes and model predictions after attaching morpholine to four different scaffolds. Ground-truth (GT) values are from ChEMBL. dir denotes the predicted direction of change; \textcolor{blue}{blue} and \textcolor{red}{red} indicate correct and incorrect dir predictions. Details of the used molecules are provided in Table~\ref{app:selfies}.}
\label{fig:context_dependent}
\end{table}

\begin{table}[t]
\centering
\footnotesize
\begin{tabular}{@{}l|cc@{}}
\toprule
\textbf{Model} & Direction acc. $\uparrow$ & Conflict-pair acc. $\uparrow$ \\
\midrule
GPT-5.2 & 0.610 & 0.145 \\
\rowcolor{my}
\texttt{Ours-1B} & \textbf{0.804} & \textbf{0.582} \\
\bottomrule
\end{tabular}
\caption{Results for context-dependent substituent effect prediction. Conflict-pairs denote (substituent, axis) pairs whose effect direction flips across the two scaffolds. \textbf{Bold} indicates the best results.}
\label{tab:context928}
\end{table}

\section{Conclusion}
In this paper, we explicitly model the contribution of individual substituents to molecular properties. To this end, we introduce \texttt{MolSC}, a dataset of substituent contributions spanning three property axes, together with \texttt{MolSC-Bench}, a held-out benchmark for evaluating substituent-level reasoning. We show that existing molecular LLMs and powerful proprietary models fail to reliably predict substituent contributions, whereas training on \texttt{MolSC} substantially improves this ability and transfers to downstream tasks, including reaction prediction and molecular property regression, using only 1D molecular representations. These results establish substituent contribution learning as a key component of fine-grained molecular understanding.

\section*{Limitations}
We introduce \texttt{MolSC}, a dataset of substituent contributions that substantially improves the fine-grained molecular understanding of LLMs. Nevertheless, several limitations remain.

\paragraph{Limited property coverage.}
Molecular behavior spans a wide range of properties. In this work, \texttt{MolSC} focuses on three axes available from or derived using ChEMBL: structural-alert liability, target-specific bioactivity, and physicochemical descriptors. While these signals already support effective training and evaluation, properties beyond this scope, including those covered in MoleculeNet \citep{wu2018moleculenet}, are not captured. Expanding \texttt{MolSC} to broader property types would be an important direction for more comprehensive molecular understanding.

\paragraph{Reliance on measured properties.}
A substituent contribution is defined as the difference in a property value between a molecule and its corresponding scaffold, and can therefore only be computed when both have annotations for the same property. As such pairs are not always available, this restricts the set of molecule-scaffold pairs that \texttt{MolSC} can cover. Moreover, for the bioactivity axis, the measured differences may reflect assay-specific variability; although we compute each contribution from a pair measured within the same assay, experimental noise and assay-dependent effects may still remain. Nevertheless, even under these constraints, training on \texttt{MolSC} improves prediction on molecules, scaffolds, and substituents unseen during training (Table \ref{tab:MolSC-Bench}).

\paragraph{Context dependence of contributions.}
The contribution of a substituent is defined for each scaffold-substituent pair and may therefore vary across scaffolds. The degree of this variation is property-dependent: opposite-sign effects are frequent for some properties, whereas properties such as HBD show little directional variation. Accordingly, \texttt{MolSC} represents contributions as scaffold-conditioned property changes rather than assigning a single fixed value to each substituent.

\section*{Acknowledgements}
This work was supported by the National Research Foundation of Korea (NRF) grant funded by the Korea government (MSIT) (No.RS-2025-00517221 and No.RS-2024-00415812) and Institute of Information \& communications Technology Planning \& Evaluation (IITP) grant funded by the Korea government (MSIT) (No.RS-2024-00439328, Karma: Towards Knowledge Augmentation for Complex Reasoning (SW Starlab), No.RS-2024-00457882, AI Research Hub Project, and No.RS-2019-II190079, Artificial Intelligence Graduate School Program (Korea University)).

% Bibliography entries for the entire Anthology, followed by custom entries
%\bibliography{anthology,custom}
% Custom bibliography entries only
\bibliography{custom}

\clearpage
\appendix

\section*{Appendix}

\section{Implementation Details}
\label{app:details}

We use Llama-3.2-1B-Instruct and Llama-3.2-3B-Instruct as backbone models.\footnote{
\href{https://huggingface.co/meta-llama/Llama-3.2-1B-Instruct}{meta-llama/Llama-3.2-1B-Instruct};
\href{https://huggingface.co/meta-llama/Llama-3.2-3B-Instruct}{meta-llama/Llama-3.2-3B-Instruct}.}
All models are fine-tuned with low-rank adaptation (LoRA; \citealp{hu2022lora}) using LLaMA-Factory \citep{zheng2024llamafactory}. We use the AdamW optimizer, sequence packing for efficient training on variable-length conversations, and a cosine learning-rate schedule with linear warmup. The maximum sequence length is set to 2048 tokens, and training is performed in bf16 precision. The learning rate is set to $1\times10^{-4}$ with a warmup ratio of $0.1$. We use LoRA rank $64$ and LoRA alpha $128$. Each device processes one packed sequence per step, and gradients are accumulated over $8$ steps. All experiments are conducted on a B200 GPU.

\texttt{MolSC} fine-tuning consists of a single supervised fine-tuning stage over 149,232 instruction conversations constructed from the forward and backward templates in Section~\ref{app:examples}. We train this stage for $6$ epochs. For downstream evaluation on Mol-Instructions \citep{fang2023mol}, we further fine-tune the \texttt{MolSC}-trained checkpoint on each target task for $10$ epochs, including forward reaction prediction, retrosynthesis, reagent prediction, and property prediction. In the dual-view setting, every molecule is represented with both SMILES and SELFIES in both inputs and outputs. For molecule generation tasks, we evaluate model outputs based on the generated SELFIES strings to match the setting of existing models. For inference, we use vLLM \citep{kwon2023efficient} with greedy decoding under each model's native chat template. The same inference, parsing, and evaluation pipeline is used for all models and variants. For BBBP and BACE, we use scaffold splits following prior work. For ACNet, we use the Few subset.

\section{Additional Statistics of \texttt{MolSC}}
\label{app:stats}

We report additional statistics of \texttt{MolSC} along four aspects. Table~\ref{tab:stats_group} summarizes how substituents and contributions are distributed over scaffolds, since a scaffold must carry more than one substituent for their effects to be compared against a shared baseline. Each scaffold carries $1.81$ substituents on average but is associated with $9.61$ contributions, since each substituent is annotated on multiple property axes, and the $181{,}098$ contributions yield $961{,}864$ axis-level annotations in total. Each substituent appears on $8.82$ scaffolds on average, and $38.2\%$ of scaffolds ($38{,}262$ of $100{,}076$) carry at least two substituents while $37.7\%$ of substituents ($7{,}737$ of $20{,}541$) appear on at least two scaffolds. Table~\ref{tab:stats_delta} reports the distribution of property changes on each axis, and Table~\ref{tab:stats_diversity} reports how the contributions are distributed over individual property axes. Finally, Table~\ref{tab:crossscaffold} reports, for each axis, the standard deviation of $\Delta_P$ for the same substituent across different scaffolds together with the fraction of substituents whose effect direction reverses. Such reversals are frequent on the bioactivity axis, reaching $58.0\%$ for IC50, whereas they are rare for descriptors such as HBD, indicating that the strength of the scaffold context varies substantially across axes.

\begin{table}[h!]
\centering
\small
\resizebox{\columnwidth}{!}{%
\begin{tabular}{l|rrrrrr}
\toprule
 & mean & median & std & min & p95 & max \\
\midrule
Substituents per scaffold & 1.810 & 1 & 1.391 & 1 & 5 & 10 \\
Contributions per scaffold & 9.611 & 6 & 8.255 & 0 & 27 & 111 \\
\bottomrule
\end{tabular}
}
\caption{Distribution of substituents and contributions per scaffold in \texttt{MolSC}.}
\label{tab:stats_group}
\end{table}

\begin{table}[h!]
\centering
\small
\resizebox{\columnwidth}{!}{%
\begin{tabular}{ll|rrrr}
\toprule
\textbf{Axis} & \textbf{Sub-axis} & Mean $\Delta$ & Std & \% increase & \% decrease \\
\midrule
\multirow{5}{*}{Physicochemical}
 & PSA         & 22.98 & 22.91 & 83.2 & 16.8 \\
 & ALogP       &  0.87 &  1.03 & 78.2 & 21.8 \\
 & HBD         &  0.43 &  1.19 & 63.3 & 36.7 \\
 & QED         & $-$0.15 &  0.14 &  9.1 & 90.9 \\
 & NP-likeness & $-$0.08 &  0.36 & 40.2 & 59.8 \\
\midrule
\multirow{8}{*}{Bioactivity}
 & IC50    & 0.07 & 0.85 & 53.7 & 46.3 \\
 & Ki      & 0.11 & 0.98 & 54.2 & 45.8 \\
 & Kd      & 0.16 & 0.94 & 56.9 & 43.1 \\
 & EC50    & 0.09 & 0.91 & 54.7 & 45.3 \\
 & AC50    & 0.07 & 0.67 & 56.0 & 44.0 \\
 & XC50    & 0.12 & 0.41 & 64.5 & 35.5 \\
 & Potency & 0.03 & 0.76 & 53.7 & 46.3 \\
 & ED50    & 0.47 & 1.45 & 56.0 & 44.0 \\
\midrule
 & & & & \% introduced & \% removed \\
\cmidrule(l){5-6}
\multirow{5}{*}{Structural-alert}
 & PAINS  & --- & --- & 100.0 &  0.0 \\
 & BMS    & --- & --- &  88.6 & 11.4 \\
 & MLSMR  & --- & --- &  96.0 &  4.0 \\
 & Dundee & --- & --- &  88.5 & 11.5 \\
 & Glaxo  & --- & --- &  77.4 & 22.6 \\
\bottomrule
\end{tabular}%
}
\caption{Distribution of property changes in \texttt{MolSC} by axis. Structural alerts are binary and are therefore reported as the fraction of changes that introduce or remove an alert.}
\label{tab:stats_delta}
\end{table}

\begin{table}[h!]
\centering
\footnotesize
\resizebox{\columnwidth}{!}{%
\begin{tabular}{ll|rr}
\toprule
\textbf{Axis} & \textbf{Sub-axis} & $n$ & Ratio (\%) \\
\midrule
\textbf{Physicochemical} & & \textbf{728{,}054} & \textbf{75.7} \\
 & ALogP       & 179{,}768 & 18.7 \\
 & QED         & 177{,}893 & 18.5 \\
 & NP-likeness & 177{,}598 & 18.5 \\
 & PSA         & 131{,}952 & 13.7 \\
 & HBD         &  60{,}843 &  6.3 \\
\midrule
\textbf{Structural-alert} & & \textbf{74{,}131} & \textbf{7.7} \\
 & MLSMR  & 30{,}383 & 3.2 \\
 & Dundee & 28{,}120 & 2.9 \\
 & BMS    &  7{,}906 & 0.8 \\
 & Glaxo  &  6{,}711 & 0.7 \\
 & PAINS  &  1{,}011 & 0.1 \\
\midrule
\textbf{Bioactivity} & & \textbf{159{,}679} & \textbf{16.6} \\
 & IC50    & 102{,}094 & 10.6 \\
 & Ki      &  24{,}409 &  2.5 \\
 & Potency &  16{,}786 &  1.7 \\
 & EC50    &  12{,}424 &  1.3 \\
 & AC50    &   1{,}811 &  0.2 \\
 & Kd      &   1{,}791 &  0.2 \\
 & XC50    &     248 &  0.0 \\
 & ED50    &     116 &  0.0 \\
\midrule
\textbf{Total} & & \textbf{961{,}864} & \textbf{100.0} \\
\bottomrule
\end{tabular}
}
\caption{Number of axis-level annotations in \texttt{MolSC} by property axis.}
\label{tab:stats_diversity}
\end{table}

\begin{table}[ht]
\centering
\small
\resizebox{\columnwidth}{!}{%
\begin{tabular}{ll|rrr}
\toprule
\textbf{Axis} & \textbf{Sub-axis} & $\Delta_P$ std & Opp.-sign (\%) & \# eligible \\
\midrule
\multirow{5}{*}{Physicochemical}
 & PSA         & 1.382 &  4.2 & 7{,}424 \\
 & ALogP       & 0.087 &  4.6 & 7{,}657 \\
 & HBD         & 0.025 &  0.0 & 4{,}474 \\
 & QED         & 0.081 & 25.8 & 7{,}642 \\
 & NP-likeness & 0.210 & 46.6 & 7{,}617 \\
\midrule
\multirow{8}{*}{Bioactivity}
 & IC50    & 0.463 & 58.0 & 13{,}385 \\
 & Ki      & 0.497 & 54.2 &  3{,}424 \\
 & Kd      & 0.380 & 45.3 &    285 \\
 & EC50    & 0.481 & 55.1 &  1{,}805 \\
 & AC50    & 0.410 & 72.1 &    154 \\
 & XC50    & 0.295 & 70.0 &     20 \\
 & Potency & 0.478 & 70.1 &  1{,}517 \\
 & ED50    & 0.495 & 47.6 &     21 \\
\midrule
\multirow{5}{*}{Structural-alert}
 & PAINS  & 0.005 &  1.6 & 7{,}737 \\
 & BMS    & 0.036 &  9.9 & 7{,}737 \\
 & MLSMR  & 0.147 & 33.9 & 7{,}737 \\
 & Dundee & 0.136 & 32.0 & 7{,}737 \\
 & Glaxo  & 0.033 &  8.9 & 7{,}737 \\
\bottomrule
\end{tabular}
}
\caption{Variance of $\Delta_P$ and frequency of opposite-sign effects for the same substituent across different scaffolds. Eligible substituents are those appearing on at least two scaffolds with a defined effect on the given axis.}
\label{tab:crossscaffold}
\end{table}

\section{Backbone Performance on Downstream Tasks}
\label{app:backbone}

\begin{table*}[h]
\centering
\resizebox{\textwidth}{!}{%
\begin{tabular}{@{}l|l|ccccccc@{}}
\toprule
\textbf{Task} & \textbf{Model} 
& Exact$\uparrow$ & BLEU$\uparrow$ & Lev$\downarrow$ 
& RDK$\uparrow$ & MAC$\uparrow$ & Mor$\uparrow$ & Validity$\uparrow$ \\ 
\midrule

Forward Reaction Prediction 
& Llama-3.2-1B-Instruct 
& 0.778 & 0.986 & 4.993 & 0.906 & 0.950 & 0.889 & 1.000 \\

Retrosynthesis 
& Llama-3.2-1B-Instruct 
& 0.574 & 0.962 & 8.707 & 0.858 & 0.909 & 0.825 & 1.000 \\

Reagent Prediction 
& Llama-3.2-1B-Instruct 
& 0.206 & 0.685 & 16.196 & 0.480 & 0.587 & 0.466 & 1.000 \\

\bottomrule
\end{tabular}%
}
\caption{Results for reaction prediction tasks of our backbone model.}
\label{tab:backbone_reaction}
\end{table*}

\begin{table}[ht!]
\footnotesize
\centering
\resizebox{\columnwidth}{!}{%
\begin{tabular}{@{}l|cccc@{}}
\toprule
\textbf{Model} & HOMO$\downarrow$ & LUMO$\downarrow$ & $\Delta\epsilon\downarrow$ & Avg$\downarrow$ \\ \midrule
Llama-3.2-1B-Instruct  & 0.0036 & 0.0038 & 0.0045 & 0.0040 \\
\bottomrule
\end{tabular}%
}
\caption{Results for property regression tasks of our backbone model.}
\label{tab:backbone_property}
\end{table}

This section reports the performance of the Llama-3.2-1B-Instruct backbone under our experimental setting. On \texttt{MolSC-Bench}, the backbone achieves an SR of $0$ on both Task~1 and Task~2, indicating that it cannot produce correct responses. As shown in Table~\ref{tab:backbone_property}, it also underperforms our fine-tuned variants on property regression. For reaction prediction tasks, Table~\ref{tab:backbone_reaction} shows that property profiling alone can degrade performance relative to the backbone, whereas adding substituent contribution supervision improves performance. These results further support the importance of explicitly learning substituent-level effects.

\section{Effect of Property Change Supervision}
\label{app:nodeltap}

Each instruction conversation in \texttt{MolSC} supervises three abilities at once (\S\ref{method:data}): property profiling, structural synthesis or decomposition, and prediction of the property changes induced by a substituent. Since the second of these is structurally similar to reaction prediction, we isolate the third by training a variant that keeps the same conversations but removes the property-change targets, so that the model performs only synthesis and decomposition. As shown in Table~\ref{tab:nodeltap}, adding the property-change targets consistently improves both Exact Match and Levenshtein distance on forward reaction prediction and retrosynthesis. This indicates that structural synthesis and decomposition alone do not account for the gains on reaction tasks, and that supervising $\Delta_P$ provides an additional contribution.

\begin{table}[!ht]
\centering
\footnotesize
% \resizebox{\columnwidth}{!}{%
\begin{tabular}{l|cc|cc}
\toprule
\multirow{2}{*}{\textbf{Method}} & \multicolumn{2}{c|}{\textbf{Forward}} & \multicolumn{2}{c}{\textbf{Retro}} \\
\cmidrule(l){2-3} \cmidrule(l){4-5}
 & Exact$\uparrow$ & Lev$\downarrow$ & Exact$\uparrow$ & Lev$\downarrow$ \\
\midrule
Ours-1B w/o $\Delta_P$ & 0.917 & 1.333 & 0.658 & 6.192 \\
\rowcolor{my}
Ours-1B & \textbf{0.927} & \textbf{1.281} & \textbf{0.674} & \textbf{5.588} \\
\bottomrule
\end{tabular}
% }
\caption{Effect of removing the property-change targets from the \texttt{MolSC} conversations. \textbf{Bold} indicates the best results.}
\label{tab:nodeltap}
\end{table}

\section{Comparing our Model with Pretrained Molecule Models}
\label{app:pretrained}

To examine how our approach compares with models that rely on large-scale molecular pretraining rather than instruction tuning, we evaluate two groups of such models. For the reaction tasks, we compare with Text+Chem T5 \citep{christofidellis2023unifying} and BioT5+ \citep{pei2024biot5+}, two small language models pretrained on large molecular corpora. For property regression, we fine-tune two pretrained molecular encoders, MoLFormer-XL \citep{ross2022large} and MolBridge \citep{park2025bridging}, on the same training data used in our experimental setting.

As shown in Table~\ref{tab:small_lm_reaction}, our models outperform both small language models on forward reaction prediction and retrosynthesis. Notably, Ours-1B improves over BioT5+ by $0.063$ and $0.032$ Exact Match on the two tasks, although BioT5+, which is pretrained on 2M PubChem and 2M PubMed records, remains competitive on reagent prediction. Our models also outperform both pretrained encoders on the property regression tasks (Table~\ref{tab:encoder_property}). These results demonstrate that \texttt{MolSC} narrows the performance gap between large-scale molecular pretraining and LLM-based methods.

\begin{table*}[h!]
\centering
\resizebox{\textwidth}{!}{%
\begin{tabular}{l|c|c|ccccccc}
\toprule
\textbf{Method} & \# Params & Modality & Exact$\uparrow$ & BLEU$\uparrow$ & Lev$\downarrow$ & RDK$\uparrow$ & MAC$\uparrow$ & Mor$\uparrow$ & Validity$\uparrow$ \\
\midrule
\rowcolor{gray}
\multicolumn{10}{l}{\textit{Forward Reaction Prediction}} \\
Text+Chem T5 \citep{christofidellis2023unifying} & 223M & 1D & 0.239 & 0.782 & 20.413 & 0.705 & 0.789 & 0.652 & 0.762 \\
BioT5+ \citep{pei2024biot5+} & 252M & 1D & 0.864 & \uline{0.993} & 3.403 & 0.949 & 0.975 & 0.935 & 1.000 \\
\rowcolor{my}
Ours-1B & 1B & 1D & \uline{0.927} & 0.990 & \uline{1.281} & \uline{0.977} & \uline{0.986} & \uline{0.969} & 1.000 \\
\rowcolor{my}
Ours-3B & 3B & 1D & \textbf{0.948} & \textbf{0.995} & \textbf{0.760} & \textbf{0.984} & \textbf{0.992} & \textbf{0.979} & 1.000 \\
\midrule

\rowcolor{gray}
\multicolumn{10}{l}{\textit{Retrosynthesis}} \\
Text+Chem T5 \citep{christofidellis2023unifying} & 223M & 1D & 0.141 & 0.765 & 24.043 & 0.685 & 0.765 & 0.585 & 0.698 \\
BioT5+ \citep{pei2024biot5+} & 252M & 1D & 0.642 & 0.969 & 6.710 & 0.897 & 0.930 & 0.866 & 1.000 \\
\rowcolor{my}
Ours-1B & 1B & 1D & \uline{0.674} & \uline{0.971} & \uline{5.588} & \uline{0.915} & \uline{0.940} & \uline{0.885} & 1.000 \\
\rowcolor{my}
Ours-3B & 3B & 1D & \textbf{0.692} & \textbf{0.972} & \textbf{5.385} & \textbf{0.920} & \textbf{0.943} & \textbf{0.894} & 1.000 \\
\midrule

\rowcolor{gray}
\multicolumn{10}{l}{\textit{Reagent Prediction}} \\
Text+Chem T5 \citep{christofidellis2023unifying} & 223M & 1D & 0.000 & 0.225 & 49.323 & 0.039 & 0.186 & 0.052 & 0.313 \\
BioT5+ \citep{pei2024biot5+} & 252M & 1D & \uline{0.257} & 0.695 & \textbf{12.901} & \uline{0.539} & \uline{0.621} & \uline{0.512} & 1.000 \\
\rowcolor{my}
Ours-1B & 1B & 1D & 0.235 & 0.732 & 14.274 & 0.516 & 0.612 & 0.495 & 1.000 \\
\rowcolor{my}
Ours-3B & 3B & 1D & \textbf{0.262} & \textbf{0.747} & \uline{13.456} & \textbf{0.548} & \textbf{0.636} & \textbf{0.529} & 1.000 \\
\bottomrule
\end{tabular}%
}
\caption{Comparison with small language models pretrained on large-scale molecular data on the reaction prediction tasks. \textbf{Bold} and \uline{underlined} mark the best and second-best scores within each task. }
\label{tab:small_lm_reaction}
\end{table*}

\begin{table*}[ht!]
\centering
\small
\begin{tabular}{l|c|c|cccc}
\toprule
\textbf{Method} & \# Params & Modality & HOMO$\downarrow$ & LUMO$\downarrow$ & $\Delta\epsilon\downarrow$ & Avg$\downarrow$ \\
\midrule
MoLFormer-XL \citep{ross2022large} & 48M & 1D & 0.0061 & 0.0063 & 0.0094 & 0.0073 \\
MolBridge \citep{park2025bridging} & 48M & 1D & 0.0064 & 0.0058 & 0.0087 & 0.0070 \\
\rowcolor{my}
Ours-1B & 1B & 1D & \uline{0.0033} & \textbf{0.0030} & \uline{0.0045} & \uline{0.0034} \\
\rowcolor{my}
Ours-3B & 3B & 1D & \textbf{0.0029} & \uline{0.0031} & \textbf{0.0039} & \textbf{0.0033} \\
\bottomrule
\end{tabular}
\caption{Comparison with pretrained molecular encoders on the QM9 property regression tasks, fine-tuned on the same training data as our experimental setting. \textbf{Bold} and \uline{underline} indicate the best and second-best results.}
\label{tab:encoder_property}
\end{table*}

\section{Results across Random Seeds}
\label{app:seeds}

To verify that the differences between our ablation settings are stable, we repeat the ablation with three random seeds on reagent prediction and QM9 property regression, the two tasks on which the settings differ by the smallest margins. As shown in Tables~\ref{tab:seeds_reagent} and \ref{tab:seeds_qm9}, the effect of dual-view representation learning is consistent across seeds: it improves the QM9 average MAE under the w/o SC setting ($p = 0.010$) and reagent prediction Exact Match under the w/ SC setting ($p = 0.006$), both by Welch's $t$-test.

\begin{table*}[h!]
\centering
\small
\resizebox{\linewidth}{!}{%
\begin{tabular}{l|cccccc}
\toprule
\textbf{Method} & Exact$\uparrow$ & BLEU$\uparrow$ & Lev$\downarrow$ & RDK$\uparrow$ & MAC$\uparrow$ & Mor$\uparrow$ \\
\midrule
Ours-1B w/o SC, w/o dual-view & 0.201 $\pm$ 0.009 & 0.690 $\pm$ 0.018 & 16.36 $\pm$ 0.46 & 0.484 $\pm$ 0.008 & 0.584 $\pm$ 0.006 & 0.458 $\pm$ 0.005 \\
Ours-1B w/o dual-view & 0.206 $\pm$ 0.005 & 0.702 $\pm$ 0.006 & 15.75 $\pm$ 0.24 & 0.497 $\pm$ 0.006 & 0.592 $\pm$ 0.002 & 0.469 $\pm$ 0.004 \\
Ours-1B w/o SC & 0.230 $\pm$ 0.002 & 0.726 $\pm$ 0.008 & 14.61 $\pm$ 0.28 & 0.506 $\pm$ 0.011 & 0.605 $\pm$ 0.006 & 0.489 $\pm$ 0.007 \\
\rowcolor{my}
Ours-1B & \textbf{0.235} $\pm$ 0.007 & \textbf{0.732} $\pm$ 0.004 & \textbf{14.27} $\pm$ 0.12 & \textbf{0.516} $\pm$ 0.002 & \textbf{0.612} $\pm$ 0.001 & \textbf{0.495} $\pm$ 0.001 \\
\bottomrule
\end{tabular}%
}
\caption{Results for reagent prediction across three random seeds, reported as mean $\pm$ standard deviation. \textbf{Bold} indicates the best results.}
\label{tab:seeds_reagent}
\end{table*}

\begin{table*}[h!]
\centering
\small
\resizebox{\linewidth}{!}{%
\begin{tabular}{l|cccc}
\toprule
\textbf{Method} & HOMO$\downarrow$ & LUMO$\downarrow$ & $\Delta\epsilon\downarrow$ & Avg$\downarrow$ \\
\midrule
Ours w/o SC, w/o dual-view & 0.00333 $\pm$ 0.00009 & 0.00380 $\pm$ 0.00033 & 0.00400 $\pm$ 0.00012 & 0.00370 $\pm$ 0.00015 \\
Ours w/o dual-view & 0.00340 $\pm$ 0.00002 & 0.00350 $\pm$ 0.00008 & 0.00402 $\pm$ 0.00030 & 0.00361 $\pm$ 0.00013 \\
Ours w/o SC & \textbf{0.00271} $\pm$ 0.00002 & \textbf{0.00273} $\pm$ 0.00003 & \textbf{0.00322} $\pm$ 0.00007 & \textbf{0.00287} $\pm$ 0.00002 \\
\rowcolor{my}
Ours-1B & 0.00276 $\pm$ 0.00003 & 0.00299 $\pm$ 0.00035 & 0.00335 $\pm$ 0.00016 & 0.00303 $\pm$ 0.00016 \\
\bottomrule
\end{tabular}%
}
\caption{Results for QM9 property regression across three random seeds, reported as mean $\pm$ standard deviation. \textbf{Bold} indicates the best results.}
\label{tab:seeds_qm9}
\end{table*}

\section{Training Data Scale of Baselines}
\label{app:budget}

All baselines compared in Sections~\ref{molecule generation} and \ref{property} undergo an additional stage of training on molecular data before downstream fine-tuning, as summarized in Table~\ref{tab:budget}. \texttt{MolSC} is comparable to or smaller in scale than the data used in these stages, and our models reach higher performance than baselines that use both more molecular data and larger backbones.

\begin{table*}[h!]
\centering
\small
% \resizebox{\columnwidth}{!}{%
\begin{tabular}{l|c|lr}
\toprule
\textbf{Model} & \# Params & \textbf{Pre-downstream stage} & \textbf{Data scale} \\
\midrule
InstructMol \citep{instructmol} & 7B & Pretraining & 264K \\
HIGHT \citep{chenhierarchical} & 7B & Pretraining & 295K \\
UniMoT \citep{guo2025unified} & 7B & Pretraining & 324K \\
Omni-Mol \citep{huomni} & 2B & Pretraining & 310K \\
KnowMol \citep{yang2025knowmol} & 7B & Instruction tuning & 100K \\
\rowcolor{my}
Ours & 1B & Instruction tuning & 149K \\
\bottomrule
\end{tabular}
% }
\caption{Training stage and data scale used by each baseline prior to downstream fine-tuning.}
\label{tab:budget}
\end{table*}

\section{Examples \& Templates}
\label{app:examples}

\begin{figure*}[hb!]
\begin{tcolorbox}[fontupper=\small\ttfamily,
                  title=Forward conversation]
\textbf{T0 [human]}\\
Tell me about physicochemical properties, structural alert flags, and target activities.\\[6pt]
\hspace*{1em}[SELFIES]\\
\hspace*{1em}\seqsplit{[C][C][C][C][C][C][Branch1][C][O][C][\#C][C][\#C][C][Branch1][C][O][C][C][C]}\\

\textbf{T1 [gpt]}\\
The molecule has the following profile.
Properties: psa=40.460, alogp=2.100, hba=2, hbd=2, mw=222.330, qed=0.530,
ro5\_viol=0, aromatic\_rings=0, heavy\_atoms=16, np\_likeness=1.600, rtb=6,
ro3\_pass=False.
Alerts: PAINS=no, Glaxo=no, BMS=no, Dundee=yes, MLSMR=yes.
Bioactivity: IC50 on ADMET pchembl=4.72.\\[6pt]

\textbf{T2 [human]}\\
Apply substituent\\
\hspace*{1em}[SELFIES]\\
\hspace*{1em}\seqsplit{[C][C][C][C]}\\
to the scaffold and tell me the resulting combined molecule.
For the affected property axes (alogp, np\_likeness, qed) and bioactivity
stypes (IC50), report the signed change as combined $-$ scaffold (positive
means the fragment increased the axis).  For every alert set
(PAINS, Glaxo, BMS, Dundee, MLSMR), state whether it was introduced, removed,
or preserved.\\[6pt]

\textbf{T3 [gpt]}\\
Combined molecule:\\
\hspace*{1em}[SELFIES]\\
\hspace*{1em}\seqsplit{[C][C][C][C][C][C][C][C][C][C][Branch1][C][O][C][\#C][C][\#C][C][Branch1][C][O][C][C][C]}\\
Effect:\\
\hspace*{1em}alogp: $+$1.560\\
\hspace*{1em}qed: $-$0.060\\
\hspace*{1em}np\_likeness: $-$0.260\\
\hspace*{1em}IC50 on ADMET: $-$0.20\\
\hspace*{1em}PAINS: preserved\\
\hspace*{1em}Glaxo: introduced\\
\hspace*{1em}BMS: introduced\\
\hspace*{1em}Dundee: preserved\\
\hspace*{1em}MLSMR: preserved
\end{tcolorbox}
\caption{A SFT example for the forward direction: starting from a scaffold profile, the model is then asked to attach a substituent and report the resulting original (combined) molecule together with the induced property changes. \texttt{stype} is short for standard type.}
\label{appendix:sft-forward}
\end{figure*}

\begin{figure*}[h!]
\begin{tcolorbox}[fontupper=\small\ttfamily,
                  title=Backward conversation]
\textbf{T0 [human]}\\
Given\\
\hspace*{1em}[SELFIES]\\
\hspace*{1em}\seqsplit{[C][=C][C][C][Branch1][=Branch1][C][C][C][C][C][N][C][Ring1][\#Branch2][=O]}\\
output its full property, alert, and bioactivity profile.\\[6pt]

\textbf{T1 [gpt]}\\
The molecule has the following profile.
Properties: psa=29.100, alogp=2.010, hba=1, hbd=1, mw=167.250, qed=0.500,
ro5\_viol=0, aromatic\_rings=0, heavy\_atoms=12, np\_likeness=1.370, rtb=4,
ro3\_pass=False.
Alerts: PAINS=no, Glaxo=no, BMS=yes, Dundee=yes, MLSMR=yes.
Bioactivity: IC50 on L1210 (Mus musculus) pchembl=4.23;
IC50 on NALM-6 (Homo sapiens) pchembl=4.08.\\[6pt]

\textbf{T2 [human]}\\
Find a scaffold + fragment decomposition for this molecule and tell me what
each piece is.  For the affected property axes (alogp, np\_likeness, qed) and
bioactivity stypes (IC50), report the signed change as combined $-$ scaffold
(positive = the fragment increased that axis).  For every alert set
(PAINS, Glaxo, BMS, Dundee, MLSMR), state whether it was introduced, removed,
or preserved.\\[6pt]

\textbf{T3 [gpt]}\\
Scaffold:\\
\hspace*{1em}[SELFIES]\\
\hspace*{1em}\seqsplit{[C][=C][C][C][Branch1][C][C][N][C][Ring1][=Branch1][=O]}\\
\hspace*{0.5em}Fragment:\\
\hspace*{1em}[SELFIES]\\
\hspace*{1em}\seqsplit{[C][C][C][C]}\\
Effect:\\
\hspace*{1em}alogp: $+$1.560\\
\hspace*{1em}qed: $+$0.050\\
\hspace*{1em}np\_likeness: $+$0.200\\
\hspace*{1em}IC50 on L1210 (Mus musculus): $-$0.47\\
\hspace*{1em}PAINS: preserved\\
\hspace*{1em}Glaxo: preserved\\
\hspace*{1em}BMS: preserved\\
\hspace*{1em}Dundee: preserved\\
\hspace*{1em}MLSMR: preserved
\end{tcolorbox}
\caption{A SFT example for the backward direction: starting from a profile of the original molecule, the model is then asked to decompose it into a scaffold and a substituent and report the property changes attributable to the substituent.}
\label{appendix:sft-backward}
\end{figure*}

\begin{figure*}[h!]
\begin{tcolorbox}[fontupper=\footnotesize\ttfamily,
                  title=Task~1 prompt]
Predict the molecular properties of the following molecule.\\[2pt]
Molecule (SELFIES):\\
\hspace*{1em}\seqsplit{[C][N][Branch2][Ring1][=Branch1][C][C][C][C][C][N][Ring1][Branch1][C][=N][C][=N][C][NH1][C][=C][C][Ring1][=Branch2][=Ring1][Branch1][C][=Branch1][C][=O][C][C][\#N]}\\[4pt]
STRICT OUTPUT FORMAT --- no reasoning, no commentary, no markdown, no preamble.\\
Respond with ONLY the lines below, one per line:\\
\hspace*{1em}<name>=<value>\\[4pt]
Properties to predict (12 numeric): psa, alogp, hba, hbd, mw, qed,
ro5\_viol, aromatic\_rings, heavy\_atoms, np\_likeness, rtb, ro3\_pass\\
Alerts to predict (5, yes/no): PAINS, Glaxo, BMS, Dundee, MLSMR\\
Bioactivity targets to predict (output pchembl value):\\
\hspace*{1em}- IC50 on Tyrosine-protein kinase JAK3 (Homo sapiens)\\[6pt]

\rule{\linewidth}{0.4pt}\\[4pt]
\textbf{Ground truth.}\\
Properties: psa=88.91, alogp=1.30, hba=5, hbd=1, mw=298.35, qed=0.92,
ro5\_viol=0, aromatic\_rings=2, heavy\_atoms=22, np\_likeness=$-$1.25,
rtb=4, ro3\_pass=no.\\
Alerts: PAINS=no, Glaxo=no, BMS=no, Dundee=no, MLSMR=yes.\\
Bioactivity: IC50 on Tyrosine-protein kinase JAK3 (Homo sapiens) pchembl=6.18.
\end{tcolorbox}
\caption{An example of a \texttt{MolSC-Bench} Task~1 prompt used for existing model inference, with the ground-truth profile shown below it.}
\label{appendix:test-task1}
\end{figure*}

\begin{figure*}[h!]
\begin{tcolorbox}[fontupper=\footnotesize\ttfamily,
                  title=Task~2 prompt]
A scaffold molecule (SELFIES):\\
\hspace*{1em}\seqsplit{[C][C][C][C][C][N][Ring1][Branch1][C][=N][C][=N][C][NH1][C][=C][C][Ring1][=Branch2][=Ring1][Branch1]}\\[4pt]
This scaffold has the following axis values:\\
\hspace*{1em}- psa = 44.81\\
\hspace*{1em}- alogp = 1.95\\
\hspace*{1em}- qed = 0.77\\
\hspace*{1em}- np\_likeness = $-$0.95\\
\hspace*{1em}- MLSMR = no\\
\hspace*{1em}- IC50 on Tyrosine-protein kinase JAK3 (Homo sapiens) = 7.09\\[4pt]
A substituent fragment (SELFIES):\\
\hspace*{1em}\seqsplit{[C][N][C][=Branch1][C][=O][C][C][\#N]}\\
is attached to the scaffold to form a new molecule.\\[4pt]
Predict the new values of the SAME axes for the resulting molecule.\\
STRICT OUTPUT FORMAT --- no reasoning, no commentary, no markdown, no preamble.\\
Respond with ONLY one line per axis in this exact format:\\
\hspace*{1em}<axis>=<value>\\[4pt]
Numeric axes: output a number.
Alert axes (PAINS, Glaxo, BMS, Dundee, MLSMR): output yes or no.\\[4pt]
Axes to predict:\\
\hspace*{1em}- psa\\
\hspace*{1em}- alogp\\
\hspace*{1em}- qed\\
\hspace*{1em}- np\_likeness\\
\hspace*{1em}- MLSMR\\
\hspace*{1em}- IC50 on Tyrosine-protein kinase JAK3 (Homo sapiens)\\[6pt]

\rule{\linewidth}{0.4pt}\\[4pt]
\textbf{Ground truth (signed contribution = combined $-$ scaffold).}\\
psa            : 44.81 $\to$ 88.91 ($\Delta$=$+$44.10, increase)\\
alogp          :  1.95 $\to$  1.30 ($\Delta$=$-$0.65, decrease)\\
qed            :  0.77 $\to$  0.92 ($\Delta$=$+$0.15, increase)\\
np\_likeness   : $-$0.95 $\to$ $-$1.25 ($\Delta$=$-$0.30, decrease)\\
MLSMR (alert)  : no $\to$ yes (introduced)\\
IC50 on JAK3   :  7.09 $\to$  6.18 ($\Delta$=$-$0.91, decrease)
\end{tcolorbox}
\caption{An example of a \texttt{MolSC-Bench} Task~2 prompt used for existing model inference (e.g., GPT-5.2, Gemini-3-Flash), with the ground-truth profile shown below it.}
\label{appendix:test-task2}
\end{figure*}

\clearpage

\begin{table*}[h!]
\centering\footnotesize
\begin{tabular}{@{} m{0.2\linewidth} m{0.78\linewidth} @{}}
\toprule
\textbf{Turn} & \textbf{Paraphrasings}\\
\midrule

\raggedright T0 (human)\newline Profile request
& 
  \begin{enumerate}[leftmargin=*,nosep,label=(\arabic*)]
    \item Predict the full molecular properties, structural alerts, and bioactivity of this molecule: \texttt{\{mol\}}.
    \item What are the molecular properties, structural alerts, and target bioactivity of \texttt{\{mol\}}?
    \item For the molecule \texttt{\{mol\}}, give a complete profile of its physicochemical properties, alert hits, and bioactivity.
    \item I have a molecule: \texttt{\{mol\}}. Provide its properties, alerts, and bioactivity profile.
    \item Describe the molecular properties, structural alerts, and bioactivity of \texttt{\{mol\}}.
  \end{enumerate}\\
\midrule

\raggedright T1 (model)\newline Profile response
& 
  \begin{enumerate}[leftmargin=*,nosep,label=(\arabic*)]
    \item Properties: \texttt{\{props\}}. Structural alerts: \texttt{\{alerts\}}. Bioactivity: \texttt{\{bio\}}.
    \item Property profile - \texttt{\{props\}}. Alert flags - \texttt{\{alerts\}}. Bioactivity - \texttt{\{bio\}}.
    \item Physicochemical: \texttt{\{props\}}. Alerts: \texttt{\{alerts\}}. Activity: \texttt{\{bio\}}.
    \item Here is the full profile. Properties: \texttt{\{props\}}. Alerts: \texttt{\{alerts\}}. Bioactivity: \texttt{\{bio\}}.
    \item Computed properties: \texttt{\{props\}}. Structural alert hits: \texttt{\{alerts\}}. Measured bioactivity: \texttt{\{bio\}}.
  \end{enumerate}\\
\midrule

\raggedright T2 (human)\newline Attachment request
& 
  \begin{enumerate}[leftmargin=*,nosep,label=(\arabic*)]
    \item Apply substituent \texttt{\{sub\}} to the scaffold and tell me the resulting combined molecule. For the affected property axes (\texttt{\{affected\_props\}}) and bioactivity stypes (\texttt{\{affected\_bio\}}), report the signed change as combined $-$ scaffold. For every alert set (PAINS, Glaxo, BMS, Dundee, MLSMR), state whether it was introduced, removed, or preserved.
    \item Attach fragment \texttt{\{sub\}} to the scaffold and give me the combined molecule. For each affected property axis (\texttt{\{affected\_props\}}) and each affected bioactivity stype (\texttt{\{affected\_bio\}}), list the signed change combined $-$ scaffold. Report the 3-way direction (introduced / removed / preserved) for all five alert sets.
    \item Bond fragment \texttt{\{sub\}} onto the scaffold and write out the combined molecule. Give the signed change (combined $-$ scaffold) for: \texttt{\{affected\_props\}}. Do the same for these bioactivity stypes: \texttt{\{affected\_bio\}}. For each of the five alert sets, say introduced, removed, or preserved.
    \item Combine the scaffold with fragment \texttt{\{sub\}} and output the resulting molecule. For each affected property axis (\texttt{\{affected\_props\}}) and bioactivity stype (\texttt{\{affected\_bio\}}), report combined $-$ scaffold. Finish with the 3-way direction for all five alert sets.
    \item Take the scaffold, attach \texttt{\{sub\}}, and describe the resulting combined molecule. For the affected property axes (\texttt{\{affected\_props\}}) and bioactivity stypes (\texttt{\{affected\_bio\}}), give the signed difference combined $-$ scaffold. For every alert set, state whether the fragment introduced it, removed it, or preserved it.
  \end{enumerate}\\
\midrule

\raggedright T3 (model)\newline Attachment response
& 
  \begin{enumerate}[leftmargin=*,nosep,label=(\arabic*)]
    \item Combined molecule: \texttt{\{cm\}}. Effect: \texttt{\{chg\}}
    \item After attachment, the molecule is \texttt{\{cm\}}. Effect: \texttt{\{chg\}}
    \item The combined molecule is \texttt{\{cm\}}. Per-axis effect: \texttt{\{chg\}}
    \item Result: \texttt{\{cm\}}. Per-axis effect: \texttt{\{chg\}}
    \item Combined: \texttt{\{cm\}}. Effect: \texttt{\{chg\}}
  \end{enumerate}\\
\bottomrule
\end{tabular}
\caption{Forward direction templates: each turn lists five paraphrasings sampled uniformly at conversation construction time. Placeholders: \texttt{\{mol\}} is the anchor molecule, \texttt{\{sub\}} the substituent, \texttt{\{cm\}} the combined molecule, \texttt{\{props\}/\{alerts\}/\{bio\}} the absolute profile fields, \texttt{\{affected\_props\}/\{affected\_bio\}} comma-separated axis names only, and \texttt{\{chg\}} the signed-delta list followed by the five alert labels.}
\label{tab:templates-forward}
\end{table*}

\begin{table*}[h!]
\centering\footnotesize
\begin{tabular}{@{} m{0.2\linewidth} m{0.78\linewidth} @{}}
\toprule
\textbf{Turn} & \textbf{Paraphrasings}\\
\midrule

\raggedright T0 (human)\newline Profile request
& 
  \begin{enumerate}[leftmargin=*,nosep,label=(\arabic*)]
    \item Predict the full molecular properties, structural alerts, and bioactivity of this molecule: \texttt{\{mol\}}.
    \item What are the molecular properties, structural alerts, and target bioactivity of \texttt{\{mol\}}?
    \item For the molecule \texttt{\{mol\}}, give a complete profile of its physicochemical properties, alert hits, and bioactivity.
    \item I have a molecule: \texttt{\{mol\}}. Provide its properties, alerts, and bioactivity profile.
    \item Describe the molecular properties, structural alerts, and bioactivity of \texttt{\{mol\}}.
  \end{enumerate}\\
\midrule

\raggedright T1 (model)\newline Profile response
& 
  \begin{enumerate}[leftmargin=*,nosep,label=(\arabic*)]
    \item Properties: \texttt{\{props\}}. Structural alerts: \texttt{\{alerts\}}. Bioactivity: \texttt{\{bio\}}.
    \item Property profile --- \texttt{\{props\}}. Alert flags --- \texttt{\{alerts\}}. Bioactivity --- \texttt{\{bio\}}.
    \item Physicochemical: \texttt{\{props\}}. Alerts: \texttt{\{alerts\}}. Activity: \texttt{\{bio\}}.
    \item Here is the full profile. Properties: \texttt{\{props\}}. Alerts: \texttt{\{alerts\}}. Bioactivity: \texttt{\{bio\}}.
    \item Computed properties: \texttt{\{props\}}. Structural alert hits: \texttt{\{alerts\}}. Measured bioactivity: \texttt{\{bio\}}.
  \end{enumerate}\\
\midrule

\raggedright T2 (human)\newline Decomposition request
& 
  \begin{enumerate}[leftmargin=*,nosep,label=(\arabic*)]
    \item Find a scaffold + fragment decomposition for this molecule and tell me what each piece is. For the affected property axes (\texttt{\{affected\_props\}}) and bioactivity stypes (\texttt{\{affected\_bio\}}), report the signed change as combined $-$ scaffold. For every alert set (PAINS, Glaxo, BMS, Dundee, MLSMR), state whether it was introduced, removed, or preserved.
    \item Decompose this molecule into scaffold and fragment. List the signed change (combined $-$ scaffold) for each affected property axis (\texttt{\{affected\_props\}}) and each affected bioactivity stype (\texttt{\{affected\_bio\}}). Report the 3-way direction for all five alert sets.
    \item Break this molecule into a scaffold and a fragment. Give the signed change combined $-$ scaffold for: \texttt{\{affected\_props\}}. Do the same for these bioactivity stypes: \texttt{\{affected\_bio\}}. For each of the five alert sets, say introduced, removed, or preserved.
    \item Identify a scaffold + fragment decomposition for this molecule and write both out. For each affected property axis (\texttt{\{affected\_props\}}) and bioactivity stype (\texttt{\{affected\_bio\}}), give combined $-$ scaffold. Finish by reporting all five alert sets as introduced, removed, or preserved.
    \item Split this molecule into a scaffold and a fragment. Give the signed difference combined $-$ scaffold for the affected property axes (\texttt{\{affected\_props\}}) and the affected bioactivity stypes (\texttt{\{affected\_bio\}}). For every alert set, state whether the fragment introduced it, removed it, or preserved it.
  \end{enumerate}\\
\midrule

\raggedright T3 (model)\newline Decomposition response
& 
  \begin{enumerate}[leftmargin=*,nosep,label=(\arabic*)]
    \item Scaffold: \texttt{\{sc\}}. Fragment: \texttt{\{sub\}}. Effect: \texttt{\{chg\}}
    \item The scaffold is \texttt{\{sc\}} and the fragment is \texttt{\{sub\}}. Effect: \texttt{\{chg\}}
    \item After decomposition: scaffold \texttt{\{sc\}}, fragment \texttt{\{sub\}}. Per-axis effect: \texttt{\{chg\}}
    \item Decomposition --- scaffold: \texttt{\{sc\}}; fragment: \texttt{\{sub\}}. Effect: \texttt{\{chg\}}
    \item Decomposed: scaffold \texttt{\{sc\}}, fragment \texttt{\{sub\}}. Per-axis effect: \texttt{\{chg\}}
  \end{enumerate}\\
\bottomrule
\end{tabular}
\caption{Backward direction templates: each turn lists five paraphrasings. Placeholders follow Table~\ref{tab:templates-forward}, with \texttt{\{sc\}} denoting the scaffold (used in T3).}
\label{tab:templates-backward}
\end{table*}

\begin{table*}[p!]
\centering
\small
\begin{tabular}{@{}lllp{7cm}@{}}
\toprule
\textbf{Axis} & \textbf{Property} & \textbf{Unit / Range} & \textbf{Description} \\
\midrule
\multirow{12}{*}{Physicochemical descriptors}
 & PSA & \AA$^2$ & Topological polar surface area. \\
 & ALogP & log unit & Atom-based octanol-water partition coefficient. \\
 & HBD & count & Hydrogen bond donor count. \\
 & QED & [0, 1] & Quantitative estimation of drug-likeness. \\
 & NP-likeness & $\sim$[$-5$, $+5$] & Natural-product-likeness score. \\
 & HBA & count & Hydrogen bond acceptor count. \\
 & MW & g/mol & Molecular weight. \\
 & Ro5 violations & count (0-4) & Number of Lipinski rule-of-five violations. \\
 & Aromatic rings & count & Number of aromatic rings. \\
 & Heavy atoms & count & Heavy atom count. \\
 & Rotatable bonds & count & Number of rotatable bonds. \\
 & Ro3 pass & binary & Astex rule-of-three pass indicator. \\
\midrule
\multirow{5}{*}{Structural-alert liability}
 & PAINS & binary & Pan-assay interference compounds. \\
 & BMS & binary & Bristol-Myers Squibb HTS deck filter. \\
 & MLSMR & binary & NIH Molecular Libraries Small Molecule Repository excluded functionality filter. \\
 & Dundee & binary & University of Dundee NTD screening library filter. \\
 & Glaxo & binary & Glaxo Wellcome hard filter. \\
\midrule
\multirow{8}{*}{Target-specific bioactivity}
 & IC50 & pchembl & Half-maximal inhibitory concentration. \\
 & Ki & pchembl & Inhibition constant. \\
 & Kd & pchembl & Dissociation constant. \\
 & EC50 & pchembl & Half-maximal effective concentration. \\
 & AC50 & pchembl & Half-maximal activity concentration. \\
 & XC50 & pchembl & Generic half-maximal concentration. \\
 & Potency & pchembl & Functional potency reported by the assay. \\
 & ED50 & pchembl & Half-maximal effective dose. \\
\bottomrule
\end{tabular}
\caption{All properties annotated in \texttt{MolSC}, organized along the three property axes. HBA, MW, Ro5 violations, aromatic rings, heavy atoms, rotatable bonds, and Ro3 pass are not considered when computing substituent contributions, since their values change monotonically under substituent attachment in our data. Bioactivity values are reported on the pchembl scale, $-\log_{10}$ of the molar concentration.}
\label{tab:property_catalog}
\end{table*}

\clearpage

\begin{table*}[ht]
\centering
\small
\renewcommand{\arraystretch}{1.8} 

\begin{tabular}{@{} >{\centering\arraybackslash}m{2cm} >{\centering\arraybackslash}m{3cm} m{0.6\linewidth} @{}}
\toprule
\textbf{Label} & \textbf{Structure} & \textbf{Textual Representation} \\
\midrule

Substituent
 & \includegraphics[width=1.6cm]{figures/substituent_morpholine.pdf}
 & \textbf{SMILES:} C1COCCN1 \newline
   \textbf{SELFIES:} \seqsplit{[C][C][O][C][C][N][Ring1][=Branch1]} \\
\midrule

Scaffold A
 & \includegraphics[width=2.3cm]{figures/scaffold_A.pdf}
 & \textbf{SMILES:} CN1Cc2ccccc2C(c2cc3ccccc3s2)C1 \newline
   \textbf{SELFIES:} \seqsplit{[C][N][C][C][=C][C][=C][C][=C][Ring1][=Branch1][C][Branch1][=C][C][=C][C][=C][C][=C][C][=C][Ring1][=Branch1][S][Ring1][=Branch2][C][Ring2][Ring1][Ring1]} \\
\midrule

Scaffold B
 & \includegraphics[width=2.3cm]{figures/scaffold_B.pdf}
 & \textbf{SMILES:} O=C(Nc1ccccc1)c1cn2ccccc2n1 \newline
   \textbf{SELFIES:} \seqsplit{[O][=C][Branch1][\#Branch2][N][C][=C][C][=C][C][=C][Ring1][=Branch1][C][=C][N][C][=C][C][=C][C][Ring1][=Branch1][=N][Ring1][=Branch2]} \\
\midrule

Scaffold C
 & \includegraphics[width=2.3cm]{figures/scaffold_C.pdf}
 & \textbf{SMILES:} O=c1c(-c2cccs2)nc2cncnc2n1-c1ccccc1 \newline
   \textbf{SELFIES:} \seqsplit{[O][=C][C][Branch1][Branch2][C][=C][C][=C][S][Ring1][Branch1][=N][C][=C][N][=C][N][=C][Ring1][=Branch1][N][Ring1][\#C][C][=C][C][=C][C][=C][Ring1][=Branch1]} \\
\midrule

Scaffold D
 & \includegraphics[width=2.3cm]{figures/scaffold_D.pdf}
 & \textbf{SMILES:} COc1ccccc1Nc1nc(Nc2ccccc2S(=O)(=O)C(C)C)c2cc[nH]c2n1 \newline
   \textbf{SELFIES:} \seqsplit{[C][O][C][=C][C][=C][C][=C][Ring1][=Branch1][N][C][=N][C][Branch2][Ring1][=Branch1][N][C][=C][C][=C][C][=C][Ring1][=Branch1][S][=Branch1][C][=O][=Branch1][C][=O][C][Branch1][C][C][C][=C][C][=C][NH1][C][Ring1][Branch1][=N][Ring2][Ring1][=Branch1]} \\

\bottomrule
\end{tabular}
\caption{Molecules used in the context-dependent contribution analysis (Table~\ref{fig:context_dependent}).}
\label{app:selfies}
\end{table*}

\end{document}